\documentclass[11pt]{article}

\usepackage[final]{acl}

\usepackage{times}
\usepackage{latexsym}

\usepackage[T1]{fontenc}

\usepackage[utf8]{inputenc}

\usepackage{microtype}

\usepackage{inconsolata}

\usepackage{graphicx}
\usepackage{subcaption}

\usepackage{url}            %
\usepackage{booktabs}       %
\usepackage{nicefrac}       %
\usepackage{xcolor}         %
\usepackage{tikz}
\usepackage{pgfplots}
\pgfplotsset{compat=1.18}
\usepackage{wrapfig}

\definecolor{bestcell}{RGB}{217,234,211}
\definecolor{worstcell}{RGB}{248,219,219}
\definecolor{winnerGreen}{HTML}{C8E6C9}
\definecolor{loserRed}{HTML}{FFCDD2}

\usepackage{amsmath}
\usepackage{amssymb}
\usepackage{mathtools}
\usepackage{amsthm}
\usepackage{comment}
\usepackage{enumitem}
\usepackage{amsfonts}

\usepackage{tcolorbox}              %
\usepackage{algorithm}
\usepackage{algorithmic}
\usepackage{adjustbox}
\usepackage{bbm}
\usepackage{pifont}
\usepackage{multirow}
\usepackage[capitalize,noabbrev]{cleveref}
\usepackage[table]{xcolor}
\usepackage{makecell}

\newcommand{\topic}[1]{\noindent\textbf{#1}}

\theoremstyle{plain}

\theoremstyle{definition}

\theoremstyle{remark}

\title{%
    Do Reasoning Representations Help Humans Evaluate LLM Outputs?
}

\author{
    Jaewoo Lim, $^{\dagger}$Sungbok Shin, and Sanghyun Hong%
    \thanks{Corresponding author.}\\
    Oregon State University, Corvallis USA \\
    $^{\dagger}$Sogang University, Seoul, South Korea \\
    \texttt{\{limjae, sanghyun.hong\}@oregonstate.edu, $^{\dagger}$sbshin90@sogang.ac.kr}
}

\usepackage{tcolorbox}
\tcbuselibrary{skins,breakable}

\providecommand{\topic}[1]{\noindent\textbf{#1}\hspace{0.4em}}

\newtcolorbox{rqbox}{
  colback=gray!8,
  colframe=black!50,
  boxrule=0.4pt,
  arc=2pt,
  left=8pt, right=8pt, top=6pt, bottom=6pt,
  before skip=8pt, after skip=8pt,
}
\begin{document}
\maketitle

\begin{abstract}
Reasoning representations are increasingly used as
explanations for large language model outputs.
Yet 
they are typically evaluated with model-centric criteria, 
such as answer accuracy and faithfulness,
leaving it unclear whether they help people evaluate model responses.
In this work, we study reasoning representations
as human-facing interfaces rather than proxies for model reasoning ability.
We conduct a controlled human study of six reasoning formats
across tasks of varying complexity,
supported by a web-based framework
that randomizes task domains, problem instances, and representation order.
The study collects fine-grained judgments of
structural understanding, error detection and localization, and trust calibration.
Our study shows a %
mismatch between 
perceived preference and support for human evaluation.
Participants prefer planning- and decomposition-based representations,
but simpler chain-of-thought traces better support 
verification, trust, and interpretability.
Preferred representations also introduce calibration risks,
with more false alarms on correct traces and 
high trust despite low willingness to verify.
\end{abstract}

\section{Introduction}
\label{sec:intro}

Large language models are increasingly used for tasks 
that require not only final answers but also reasoning traces 
that help users evaluate the outcomes.
Methods such as chain-of-thought prompting~\cite{kojima2022large, wei2022}, 
self-consistency~\cite{wang2023selfconsistency}, planning~\cite{wang2023plansolve}, 
decomposition~\cite{zhou2023leasttomost}, and template-based reasoning~\cite{yang2024buffer} 
expose such traces in different forms.
While these \emph{reasoning representations}
were originally introduced to improve model performance,
they are now used as explanations
for assessing whether a model response is correct and trustworthy.

This shift creates an evaluation mismatch.
Reasoning representations are treated as user-facing explanations,
but evaluated primarily with model-centric criteria.
Prior work asks 
whether a representation improves answer accuracy~\cite{cobbe2021gsm8k, suzgun2023bbh, sprague2024cot},
generates more faithful rationales~\cite{turpin2023language, madsen2024self}, 
or yields more coherent reasoning~\cite{golovneva2023roscoe, prasad2023receval, lee2025evaluating}.
These criteria are important yet do not establish
whether humans can use the representation to evaluate a model response.
A reasoning trace can be accurate, faithful or detailed
while still making it difficult for users 
to locate an error, understand the role of each component,
or calibrate trust in the final answer.

It becomes particularly important because
richer explanations are \emph{not}
necessarily improve human oversight~\cite{bansal2021does, bucinca2021trust}.
Structured formats,
such as plans or decomposed subproblems,
may appear more useful by organizing reasoning into explicit components.
But the same structure can 
obscure errors, misdirect attention, 
or increase trust without increasing verification.
Evaluating these representations thus
requires more than measuring user preference or model performance;
it requires measuring how they affect distinct forms of human judgment.

In this work, we shift the attention from model-side outcomes 
to human evaluation behavior in the study of reasoning representations.
Specifically, we ask: 
\emph{Do reasoning representations help humans evaluate LLM outputs?}
We focus on three evaluation behaviors 
that reasoning traces are expected to support in practice:
structural understanding, error detection and localization, and trust calibration.
This framing allows us to test whether
more structured representations %
improve human evaluation, 
or make the reasoning appear more interpretable.

To answer this question, we conduct a controlled human evaluation of reasoning representations.
We design a protocol that compares six representations
under matched task and problem conditions
across the three forms of human evaluation above.
The protocol isolates representation utility from answer correctness 
and separates perceived usefulness from verification behavior
by combining correct final-answer traces with controlled error-injected traces.
We instantiate this protocol in a web-based evaluation framework
that randomizes task domains, model families, problem instances,
error conditions, and representation order 
while collecting representation-level judgments.

Our user study with 50 participants offers 
a three-way divergence among 
preference, verification performance, and trust calibration.
(1) Participants prefer planning- and decomposition-based formats, 
suggesting that explicit structure increases perceived usefulness.
(2) However, this preference does not translate into verification performance: 
simpler chain-of-thought traces better support error detection and localization.
(3) Preferred formats also introduce calibration risks, 
such as false alarms on correct traces and high trust paired with low willingness to verify.
Those findings suggest that increasing representational structure 
does not uniformly improve human evaluation.

\smallskip
\topic{Contributions.}
Our contributions are as follows:
\begin{itemize}[
    noitemsep, 
    topsep=0.1em,
    leftmargin=1.2em]
    \item We reframe reasoning representations as interfaces for human evaluation,
    rather than solely as indicators of model reasoning ability.
    \item We design a controlled human-evaluation protocol
    for comparing reasoning representations 
    under matched tasks and problem instances.
    \item We instantiate our protocol in a web-based evaluation framework 
    that randomizes assignment, error conditions, and representation order 
    while collecting representation-level judgments.
    \item We show that reasoning representations induce distinct trade-offs 
    among preference, verification performance, and trust calibration.
\end{itemize}

\definecolor{selectedrow}{gray}{0.88}

\begin{table*}[t]
\centering
\small
\setlength{\tabcolsep}{5pt}
\renewcommand{\arraystretch}{1.05}
\resizebox{\textwidth}{!}{
\begin{tabular}{@{}r l ccc | l@{}}
\toprule
\textbf{\#} & \textbf{Method} & \textbf{Topology} & \textbf{Scope} & \textbf{Decomp.} & \textbf{Evaluation Criteria} \\
\midrule \midrule
\multicolumn{6}{@{}l}{\textit{\textbf{A. Direct prompting}\quad — single-shot generation without intermediate reasoning steps}} \\
\rowcolor{selectedrow}
1  & Standard I/O~\citep{brown2020language}                    & Chain           & SP    & Monolithic        & Task accuracy \\
\midrule
\multicolumn{6}{@{}l}{\textit{\textbf{B. Linear reasoning}\quad — single chain of explicit intermediate steps before the answer}} \\
\rowcolor{selectedrow}
2  & Zero-shot CoT~\citep{kojima2022large}                     & Chain           & SP    & Monolithic        & Task accuracy \\
3  & CoT (few-shot)~\citep{wei2022}                            & Chain           & SP    & Monolithic        & Task accuracy \\
4  & Scratchpad~\citep{nye2021show}                            & Chain           & SP    & Monolithic        & Task accuracy \\
5  & Program of Thoughts~\citep{chen2023program}               & Chain           & SP    & Monolithic        & Task accuracy \\
6  & Self-Notes~\citep{lanchantin2023learning}                 & Chain           & SP    & Monolithic        & Task accuracy \\
7  & Faithful CoT~\citep{lyu2023faithful}                      & Chain           & SP    & Monolithic        & Task accuracy, Faithfulness \\
\midrule
\multicolumn{6}{@{}l}{\textit{\textbf{C. Aggregated sampling}\quad — multiple chains sampled and combined into a single answer}} \\
\rowcolor{selectedrow}
8  & Self-Consistency CoT~\citep{wang2023selfconsistency}      & Tree$^\ddagger$ & SP+MP & Monolithic        & Task accuracy \\
9  & Skeleton-of-Thought~\citep{ning2023sot}                   & Tree$^\ddagger$ & SP+MP & Planned           & Preference / Quality, Efficiency \\ %
\midrule
\multicolumn{6}{@{}l}{\textit{\textbf{D. Planned decomposition}\quad — explicit decomposition of the problem into sub-problems before solving}} \\
\rowcolor{selectedrow}
10 & Plan-and-Solve~\citep{wang2023plansolve}                  & Chain           & SP    & Planned           & Task accuracy \\
\rowcolor{selectedrow}
11 & Least-to-Most~\citep{zhou2023leasttomost}                 & Chain           & MP    & Recursive         & Task accuracy \\
12 & Decomposed Prompting~\citep{khot2023decomposed}           & Chain           & MP    & Recursive         & Task accuracy \\
\midrule
\multicolumn{6}{@{}l}{\textit{\textbf{E. Iterative self-correction}\quad — feedback-and-refine loop over the model's own output}} \\
13 & Self-Refine~\citep{madaan2023selfrefine}                  & Chain           & MP    & Iterative         & Task accuracy, Preference / Quality \\ %
14 & Reflexion~\citep{shinn2023reflexion}                      & Chain           & MP    & Iterative         & Task accuracy \\ %
15 & Self-Debug~\citep{chen2023selfdebug}                      & Chain           & MP    & Iterative         & Task accuracy \\
16 & SelfCheck~\citep{miao2024selfcheck}                       & Chain           & MP    & Iterative         & Task accuracy, Verification performance \\ %
\midrule
\multicolumn{6}{@{}l}{\textit{\textbf{F. Multi-path search}\quad — explicit search over a tree or graph of candidate reasoning paths}} \\
17 & Tree of Thoughts~\citep{yao2023tot}                       & Tree            & MP    & Branching         & Task accuracy, Preference / Quality \\ %
18 & Algorithm of Thoughts~\citep{sel2024algorithm}            & Tree            & SP    & Branching         & Task accuracy \\
19 & Graph of Thoughts~\citep{besta2024graph}                  & Graph           & MP    & Aggregating       & 
Task accuracy, Efficiency \\
20 & Cumulative Reasoning~\citep{zhang2023cumulative}          & Graph (DAG)     & MP    & Aggregating       & Task accuracy \\
\midrule
\multicolumn{6}{@{}l}{\textit{\textbf{G. Template-grounded / meta-structure}\quad — reasoning guided by an explicit external template or self-discovered structure}} \\
\rowcolor{selectedrow}
21 & Buffer of Thoughts~\citep{yang2024buffer}                 & Chain           & MP    & Template-grounded & Task accuracy, Efficiency \\
22 & Self-Discover~\citep{zhou2024selfdiscover}                & Graph           & MP    & Meta-structure    & Task accuracy, Preference / Quality \\ %
\midrule
\multicolumn{6}{@{}l}{\textit{\textbf{H. Agentic / grounded}\quad — interleaved reasoning and actions in an external environment}} \\
23 & ReAct~\citep{yao2023react}                                & Chain           & MP    & Interleaved       & Task accuracy \\ %
\midrule
\multicolumn{6}{@{}l}{\textit{\textbf{I. Latent reasoning}\quad — reasoning carried in continuous hidden states rather than explicit tokens}} \\
24 & COCONUT~\citep{hao2024coconut}                            & Chain           & SP    & Latent            & Task accuracy, Efficiency \\
25 & Pause Tokens~\citep{goyal2024pause}                       & Chain           & SP    & Latent            & Task accuracy \\ %
\bottomrule
\end{tabular}
}
\caption{\textbf{LLM reasoning methods and their evaluation criteria.} 
We categorize 25 reasoning methods (A--I) by the structural form they expose to users. 
Structural characteristics follow~\citet{besta2024demystifying}: 
\emph{Topology} denotes chain/tree/graph; 
\emph{Scope} distinguishes single-prompt (SP) from multi-prompt (MP); and 
\emph{Decomp.} summarizes how reasoning is decomposed into intermediate steps. 
\emph{Evaluation criteria} abstracts the primary criteria used in prior work. 
Gray rows indicate the six representations evaluated in our study. 
$^\ddagger$~denotes a depth-one tree.
}
\label{tab:reasoning-representations}
\end{table*}

\section{Background and Related Work}
\label{sec:background}

\topic{Reasoning representations for LLMs.}
A \textit{reasoning trace} is the sequence of intermediate tokens 
produced before a model's final answer,
whereas a \textit{reasoning representation} is the user-visible form
in which that trace is organized and presented. %
This distinction is important because our study does not
treat reasoning traces as faithful records of a model's internal computation.
Prior work shows that CoT explanations can be unfaithful 
to the factors that actually determine model predictions~\cite{turpin2023language}. 
Our focus is therefore not whether a trace reveals how the model truly reasoned,
but whether its presentation helps humans inspect, verify, and calibrate trust
in the model output~\cite{bansal2021does,kim24llmuncertainty}.

Table~\ref{tab:reasoning-representations} 
organizes existing LLM reasoning methods
by the structural forms they expose to users.
The A--I groups reflect an overall progression in how reasoning is externalized:
from direct answering without visible intermediate reasoning~\cite{brown2020language},
to linear chains of thought~\cite{wei2022,kojima2022large},
aggregated chains~\cite{wang2023selfconsistency},
planned and decomposed reasoning~\cite{wang2023plansolve,zhou2023leasttomost,khot2023decomposed},
iterative refinement~\cite{madaan2023selfrefine,shinn2023reflexion},
tree- or graph-based search~\cite{yao2023tot,besta2024graph},
template-grounded structures~\cite{yang2024buffer,zhou2024selfdiscover},
agentic reasoning-action traces~\cite{yao2023react},
and latent reasoning~\cite{hao2024coconut,goyal2024pause}.

Across the groups, we further characterize each method 
along three structural dimensions following~\citet{besta2024demystifying}:
\emph{topology}, the organization of reasoning as a chain, tree, or graph;
\emph{scope}, whether reasoning unfolds within a single prompt-response exchange 
or across multiple controller-issued steps; and 
\emph{decomposition}, the intermediate reasoning units
exposed by the method, e.g., plans, subproblems, branches, templates, or actions.

Despite their original intent, 
reasoning outputs are increasingly repurposed 
as user-facing explanations~\cite{sun26seeingreasoning}.
This shift makes it necessary to evaluate 
not only how they perform, but also
\emph{how their representations shape human judgment}.
We use Table~\ref{tab:reasoning-representations} to identify
structurally distinct representations for controlled comparison 
(\S\ref{subsec:repr-selection}).

\smallskip
\topic{Evaluating reasoning representations.}
Most prior work on LLM reasoning evaluates methods 
through model-centric criteria, such as answer accuracy,
faithfulness, and coherence~\cite{%
    hendrycks21mathdataset, turpin2023language, lanham23faithfulness, 
    golovneva2023roscoe, prasad2023receval, lightman24stepbystep}.
While these criteria are useful for assessing reasoning quality,
they do not determine whether a representation helps users evaluate model outputs.
A reasoning trace can be accurate or faithful
while still making it difficult for users to identify
an error, understand its impact, or calibrate trust in the final answer.

Prior work on human-centered evaluation of explanations 
instead considers dimensions such as 
preference, comprehension, decision quality, and trust calibration~\cite{buccinca20proxy, bansal2021does}. 
These studies show that subjective preference 
does not necessarily translate into better behavioral outcomes 
such as error detection or decision quality, and 
that explanations can even increase reliance on incorrect outputs~\cite{kim25llmreliance}. 
This motivates evaluating perceived usefulness 
separately from behavioral outcomes 
such as error detection, localization, and trust calibration.

Despite the growing body of work, few studies 
compare structurally distinct reasoning representations under controlled conditions.
Existing work is often limited in representation coverage, 
experimental control, or behavioral evaluation~\cite{%
    si24llmstruthfulness, sun26seeingreasoning, pang26interactivereasoning}.
Moreover, most evaluations assess reasoning traces 
without measuring how users interpret, verify, or calibrate trust in them.
Our controlled human-evaluation protocol addresses this gap 
by jointly measuring structural understanding, 
error detection and localization, trust calibration,
and perceived usefulness across selected reasoning representations.

\section{Controlled Human Evaluation}
\label{sec:method}

\begin{figure*}[t]
\centering
\includegraphics[width=\textwidth]{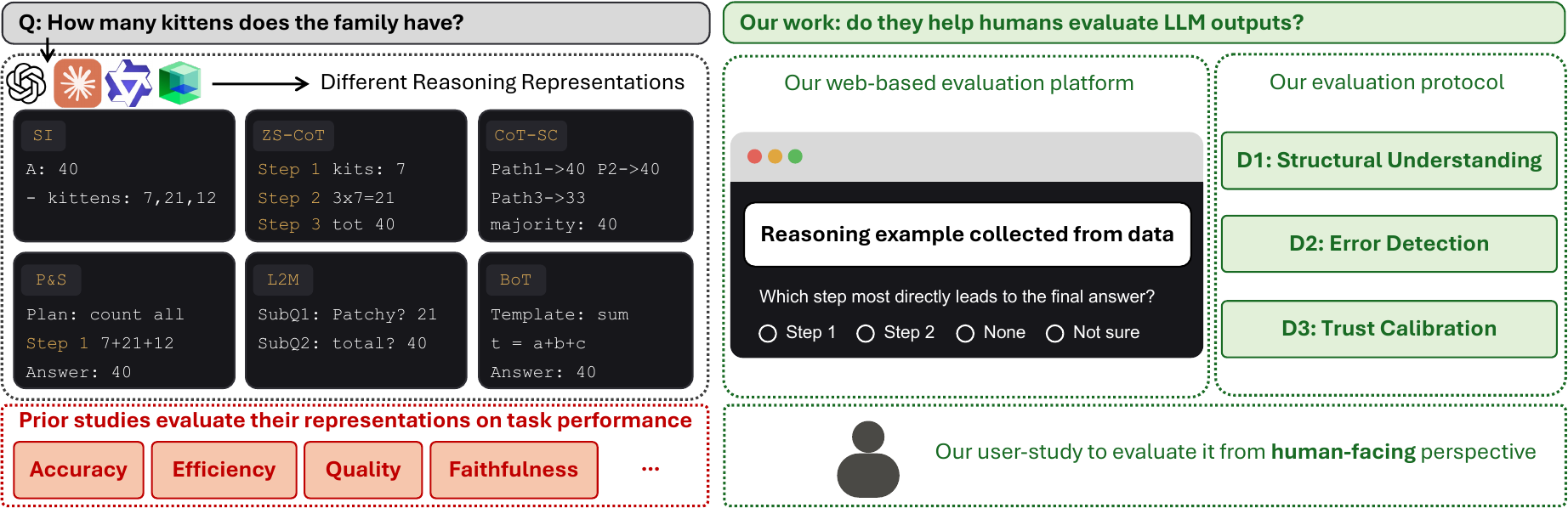}
\caption{%
\textbf{From model-centric to human-centric evaluation.} 
Reasoning representations expose the same problem 
and answer in different forms, 
e.g., a direct answer, a zero-shot CoT, 
sampled CoT chains, a plan and solution,
subproblems, or an abstract template (top-left).
These methods have been judged by model-centric metrics
(bottom-left), not by whether their traces help a user. 
Our work studies how effective they are
in helping humans evaluate model outputs:
structural understanding (D1), 
error detection and localization (D2), 
and trust calibration (D3).}

\label{fig:workflow}
\end{figure*}

We present a controlled human-evaluation protocol
for comparing reasoning representations as user-facing explanations.
Our goal is to isolate the effect of representation from final-answer correctness:
users compare traces that reach the same correct answer, 
while controlled error-injected traces provide ground truth for verification behavior.
We compare six reasoning representations 
across three benchmarks and measure three forms of human evaluation:
structural understanding, error detection and localization, and trust calibration.

\subsection{Evaluation Dimensions}
\label{sec:dimensions}

Prior work provides limited guidance 
for defining human-centered evaluation dimensions.
Studies on reasoning-trace evaluation~\cite{lee2025evaluating}
has proposed metrics for assessing trace quality, 
while human-verification studies emphasize error detection and localization%
~\cite{zhou2025icot}.
A separate line of work on explainable AI offers established constructs
for trust and reliance~\cite{jian2000foundations, hoffman2023measures}.
We adapt these strands into three human-centered dimensions:

\smallskip
\topic{D1: Structural understanding}
captures whether users can identify the role of each reasoning component---%
whether it introduces external information, performs a key inference, or marks a shift in approach%
---and trace how components depend on one another.
We measure whether users notice the structural cues
exposed by each representations, such as plan-execution, 
boundaries, subproblem transitions, or template instantiations.
These cues matter because a few steps, 
such as those that set up a plan or shift the approach,
disproportionately determine the final answer~\cite{bogdan2025thought}.
Whether a user can locate such steps depends on how the trace is presented.
This ability makes verification affordable:
explanations help only when users can actually verify the answer~\cite{fok2024verifiability},
and users verify only when the effort is low enough~\cite{vasconcelos2023explanations}.

\smallskip
\topic{D2: Error detection and localization.}
It measures whether users can detect flawed reasoning 
and localize the step or component where the error stems.
Prior work~\cite{zhou2025icot}
studies error detection as a human-verification task,
but we adapt it to compare reasoning representations 
under controlled, cross-task conditions.

\smallskip
\topic{D3: Trust calibration and reliance.}
D3 measures whether users' trust-related judgments
align with the evidence made visible by the reasoning representation.
Work on explainable AI~\cite{hoffman2023measures}
offers constructs for trust and reliance,
but we use them to examine whether preferred reasoning formats 
support verification and reliance.

\subsection{Representation Selection}
\label{subsec:repr-selection}

Guided by the taxonomy in Table~\ref{tab:reasoning-representations},
we select six reasoning representations that 
(1) cover common user-visible forms of LLM reasoning 
(2) while remaining comparable within the same evaluation protocol.
The selected formats span 
direct prompting, linear reasoning, aggregation, 
planning, decomposition, and template-grounded reasoning.
We do \emph{not} aim to exhaust the full design space;
instead, we choose structural diverse formats 
that can be rendered as step-level traces and 
evaluated using the same questions across conditions.

\smallskip
\topic{Standard I/O}~\cite{brown2020language}
produces a direct answer without unfolding reasoning trace.
This method serves as an answer-first baseline. 

\topic{Zero-shot CoT}~\cite{kojima2022large}
produces a sequence of intermediate reasoning steps,
elicited by the instruction ``let's think step by step.''
This serves as the step-by-step reasoning baseline.

\topic{Self-Consistency CoT (CoT-SC)}~\cite{wang2023selfconsistency}
samples multiple reasoning paths and aggregates final answers by agreement.
We present the majority-vote chain with its agreement score.

\topic{Plan-and-Solve}~\cite{wang2023plansolve}
first produces an explicit plan and then executes that plan step by step,
separating approach from execution.

\topic{Least-to-Most}~\cite{zhou2023leasttomost}
decomposes the problem into ordered subproblems
and solves them progressively.

\topic{Buffer of Thoughts (BoT)}~\cite{yang2024buffer}
applies a reusable thought-template
retrieved from a meta-buffer of prior solutions to the current problem,
exposing both the abstract template and its problem-specific instantiation.

\subsection{Study Materials and Sampling}
\label{sec:benchmarks}

We construct study materials
to compare reasoning representations 
across diverse reasoning demands while controlling for final-answer correctness.
We use three benchmarks:
{GSM8K}~\cite{cobbe2021gsm8k} for multi-step arithmetic reasoning,
{HotPotQA}~\cite{yang2018hotpotqa} for multi-hop factual question-answering, and
{BBH}~\cite{suzgun2023bbh} for symbolic and logical reasoning.
This selection allows us to examine
whether representation effects are specific to a task type or 
persist across qualitatively different reasoning settings.
For each benchmark, we sample candidate problems and 
generate outputs for all six representation conditions
using OpenAI's GPT-5 (\texttt{gpt-5-2025-08-07})
and the prompt instantiations described above.
Using a single model keeps model behavior fixed across conditions,
so that the comparison focuses on differences in the displayed reasoning format.
Prompt templates and generation details are provided in Appendix.

\begin{table}[ht]
\centering
\small
\adjustbox{max width=\linewidth}{
\begin{tabular}{l r c r}
\toprule
\textbf{Benchmark} & \textbf{\# Problems} & \textbf{Steps} & \textbf{Input words} \\
\midrule
GSM8K     & 9  & $4.19 \pm 0.77$ & 39.6 \\
HotPotQA  & 9  & $3.33 \pm 0.26$ & \textsuperscript{$\dagger$}156.4 \\
BBH       & 9  & $3.85 \pm 0.61$ & 89.6 \\
\midrule
All       & 27 & $3.79 \pm 0.67$ & - \\
\bottomrule
\end{tabular}
}
\caption{%
    \textbf{Problem-level statistics for the retained all-correct pool.}
    It contains 27 problems, with 9 per benchmark.
    Steps are reported as mean $\pm$ std. and 
    are averaged across the six generated traces for each problem. 
    Displayed input words count the task input shown to participants, 
    excluding prompting instructions and output-format constraints.
    \textsuperscript{$\dagger$}For HotPotQA, displayed input words 
    include the question and retrieval context.}
\label{tab:dataset_stats}
\end{table}

To isolate representation utility from answer correctness,
we retain only problem instances for which
all six representations produce the correct answer.
This all-correct filter ensures that participants 
compare traces that reach the same correct outcome, 
rather than traces that differ in final-answer accuracy.
Of 140 candidates problems, 27 (19.3\%) satisfy this filter. We resample until each benchmark has 9 problems, yielding the final pool of \emph{27 problems} with \emph{9 per benchmark}.
For each retained problem, 
we store the task input, %
the correct final answer, and six generated reasoning traces.
Table~\ref{tab:dataset_stats} summarizes the retained all-correct pool---%
the average number of reasoning steps in the generated traces 
and the amount of task input shown to participants.

\subsection{Controlled Error Injection}
\label{sec:error-injection}

To measure error detection and localization,
we require reasoning traces with known error locations.
Naturally occuring model errors are hard to control across representations,
we therefore inject controlled errors 
into a subset of otherwise correct traces.
Each injected trace contains one intended, localizable error and it is manually verified.

\smallskip
\topic{Injection targets.}
From the 27 all-correct problems,
we select 9 problems for the error-detection task (3 per benchmark).
For GSM8K and BBH, 
we inject errors into all six representations
for each selected problem.
For HotPotQA, 
errors are injected only into
Least-to-Most and Buffer of Thoughts,
whose formats support clean within-trace factual edits
without altering the retrieval context.
This yields \emph{42 errored traces}.
Because HotPotQA covers fewer representations,
we report the results both in aggregate and by benchmark,
and avoid unsupported cross-benchmark contrasts.

\topic{Injection procedure.}
Errors are created through a human-in-the-loop pipeline
and manually verified to ensure that 
each trace contains exactly one intended error.
We use two ground-truth error types adapted from prior work%
~\cite{kamoi2024evaluating}: 
\textit{incorrect calculation} (a numerical error in an arithmetic step) and 
\textit{incorrect premise} (an introduced fact or assumption 
that contradicts the problem or a prior step).
Participants first localize the error and then 
select an error type from five options: 
incorrect calculation, incorrect premise, missing reasoning step, irrelevant reasoning, and other.
The latter three options serve as distractors 
and help distinguish correct localization from generic suspicion.
Additional error-injection details are in Appendix.

\begin{figure}[!t]
\centering
\includegraphics[width=\columnwidth]{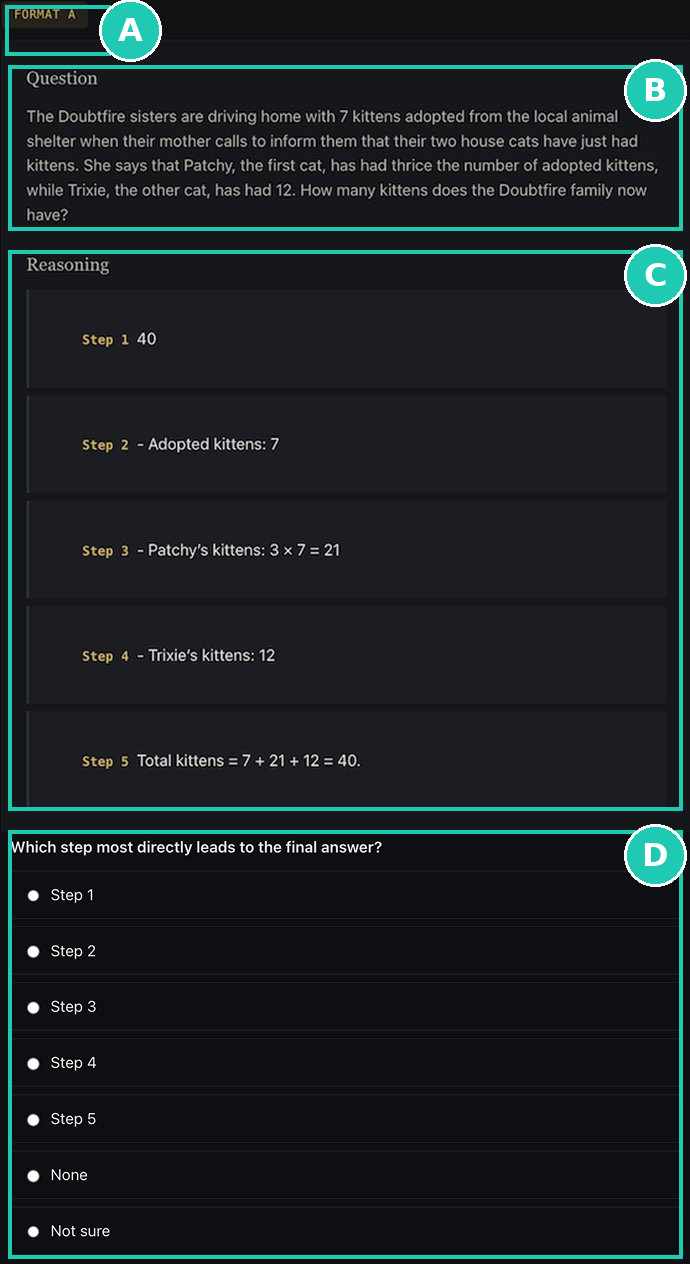}
\caption{\textbf{Web interface for the structural understanding task.} Participants view a reasoning trace rendered in one of six representations (A: representation tag; B: problem prompt; C: reasoning steps), shown alongside a dependency question (D). The example shows the key-step item; the same interface is used for the reasoning-shift and supporting-step items.}
\label{fig:reasoningvis_interface}
\end{figure}
\section{A Web-Based Evaluation Framework}
\label{sec:framework}

We instantiate the protocol in a web-based evaluation framework 
that supports controlled assignment, randomized presentation, 
branching error-detection workflows, and representation-level logging.
The framework allows participants 
to evaluate multiple reasoning representations 
under comparable task, model, problem, and error-condition settings 
while collecting fine-grained judgments for each representation.
Our protocol requires more than a static survey: 
it must assign participants to balanced study cells, 
present six representations in randomized order, 
conditionally display error-detection follow-ups, and 
record judgments at the participant-representation level.

\smallskip
\topic{Overview.}
The framework implements three functions.
(1) It assigns each participant to a benchmark, problem set, and
error-condition group.
(2) It presents the six representation conditions in randomized order for each
problem, ensuring that representation comparisons are not confounded with a fixed
presentation order.
(3) It records representation-level responses, including structural
understanding judgments, error-localization choices, trust-related ratings,
post-study preferences, and per-item timing.
The framework is deployed as a web application on Hugging Face Spaces with persistent
storage delivering the same rendered representations 
to all participants and logging response data and per-item timing for analysis.

\subsection{Participant Workflow}
\label{subsec:workflow}
Each participant completes a single online session.
Participants were recruited through in-class announcements at the authors' institution in two collection waves;
the only eligibility requirements were 
being at least 18 years of age and comfortable reading English,
with no further screening or selection.
64 participants who began the session, 51 completed all phases;
one completed after the data-collection window closed,
yielding the analytic sample of $n = 50$.
After consent and onboarding, participants complete a pre-survey 
on technical familiarity and prior experience with AI tools;
it collected background covariates only 
and was not used to exclude any participant.
The framework then assigns the participant to one benchmark 
and presents three problems from it, one per evaluation dimension,
each displaying the six representations from \S\ref{subsec:repr-selection} in randomized order.
The three problems target structural understanding (D1; participants identify reasoning shifts, key steps, and supporting information), 
error detection and localization (D2; participants evaluate a mixture of correct and error-injected traces, flag any error, and select its type when applicable), 
and trust calibration (D3; participants rate reliance, verification intention, and perceived interpretability).
The session ends with post-study comparison questions and an overall preference.

\subsection{Randomization and Assignment}
\label{subsec:assignment}

Representation order is \emph{randomized} 
within each problem to reduce order effects.
Participants are assigned across benchmark and 
error-condition groups using a balanced rotation.
For the error-detection task, 
the framework controls which representations appear in the errored condition.
For each problem, a subset of representations is shown 
with injected errors and the remaining representations 
are shown in their original correct form.
The errored conditions are rotated across participants 
so that each representation is evaluated under both correct and errored conditions 
where supported by the benchmark materials.
This assignment scheme enables representation-level comparison 
while controlling for task domain, problem instance, error condition, 
and presentation order.

\subsection{Measures and Analysis}
\label{subsec:measures}

We collect four sets of measures:

\smallskip
\topic{Structural understanding.}
Participants identify, among the steps of the displayed trace: 
(1) where the reasoning shifts to 
a different interpretation or approach (\textit{reasoning shift}); 
(2) which step most directly leads to the final answer (\textit{key step}); and
(3) which step introduces an important 
supporting fact, rule, or verification (\textit{supporting step}).
Options enumerate the steps actually present in the trace,
as well as ``None'' and ``Not sure.''
We summarize responses by representation
to examine whether participants perceive and localize structural cues.

\smallskip
\topic{Error detection and localization.}
Participants first select the step where the error occurs
(\textit{``No error''} / Step~$k$ / ``I cannot determine'').
If they identify a step, two follow-ups elicit the \textit{error type} 
and the \textit{error impact} on the final answer.
We compare participant responses against the known injected-error labels: 
detection accuracy measures
whether participants identify that a trace contains an error, and 
localization accuracy measures 
whether the selected step matches the injected-error location.

\smallskip
\topic{Trust and reliance.}
Participants answer three 5-point Likert items 
adapted from established trust measures~\cite{jian2000foundations, hoffman2023measures}:
\textit{reliance} (``To what extent would you rely on this reasoning when making a decision?'');
\textit{verification intention} (``Would you verify this reasoning before using it?''); and
\textit{interpretability} (``How easy was it to understand this reasoning?''). 
We compare response distributions 
to examine whether perceived usefulness aligns with verification behavior.

\smallskip
\topic{Post-study} collects 
comparative judgments and overall preference 
across the six representation formats.
Post-study preference items are reported 
as frequency distributions over the six representations.

\smallskip
\noindent
The full survey details are %
in Appendix~\ref{app:survey}.

\section{Evaluation}
\label{sec:results}

We evaluate six reasoning representations 
across three %
dimensions (D1--3), 
interpreting the results within each dimension 
and then across them.

\topic{Metrics.}
We report three families of metrics per representation. 
For \textit{Structural understanding} (D1), we report shift recognition (the share of responses that identify a reasoning shift), key-step, and support-step agreement, the latter two as Krippendorff's $\alpha$; we use agreement rather than dependency correctness, which is difficult to define unambiguously for human evaluation. 
Higher $\alpha$ indicates greater agreement 
among participants on which step is important, 
whereas lower $\alpha$ indicates that 
the representation does not consistently 
make the important step identifiable.
For \textit{Verification performance} (D2), we report, on 
error-injected traces, the false-alarm rate 
(flagging an error in a correct trace; lower is better) 
and localization accuracy (selecting the step 
containing the injected error; higher is better).
For \textit{Trust calibration} (D3), 
we report mean ratings on
three 5-point Likert items: 
reliance (1 = would not rely, 5 = would rely), 
verification intention (1 = definitely verify, 
5 = would not verify; lower indicates 
stronger intent to verify), 
and interpretability (1 = very difficult, 5 = very easy).
\textit{Preference} counts how often, 
out of 50 participants, each representation 
was selected on the overall 
and item-level post-survey questions.

\begin{table}[ht]
\centering
\adjustbox{max width=\linewidth}{
\begin{tabular}{lccc}
\toprule
\textbf{Representation} & 
    \makecell{\textbf{Shift}\\\textbf{Recognition}} & 
    \makecell{\textbf{Key-Step}\\\textbf{$\alpha$}} & 
    \makecell{\textbf{Support-Step}\\\textbf{$\alpha$}} \\
\midrule \midrule
Standard I/O        & 48\% & 0.34 & \textbf{0.48} \\
Zero-shot CoT       & 58\% & 0.30 & 0.07 \\
CoT-SC              & 45\% & \textbf{0.37} & 0.19 \\
Plan-and-Solve      & 60\% & 0.26 & 0.30 \\
Least-to-Most       & 56\% & 0.19 & 0.05 \\
Buffer of Thoughts  & \textbf{64\%} & 0.11 & 0.09 \\
\bottomrule
\end{tabular}
}
\caption{
\textbf{Shift Recognition} reports the percentage of responses that identified a meaningful reasoning transition rather than selecting \emph{None} or \emph{Not sure}.
\textbf{Key-Step} and \textbf{Support-Step} report inter-participant agreement as Krippendorff's $\alpha$ (nominal), computed per representation with each (problem, trace) as a unit and participants as coders; \emph{None} and \emph{Not sure} are treated as valid categories, and (participant, problem) pairs unobserved due to benchmark assignment are handled as missing data.
Higher values indicate stronger chance-corrected agreement; bold marks the highest value in each column.
}
\label{tab:rq1_summary}
\end{table}

\definecolor{winnerGreen}{HTML}{C8E6C9}
\begin{table*}[t]
\centering
\small
\setlength{\tabcolsep}{4pt}
\renewcommand{\arraystretch}{1.1}
\begin{tabular}{l@{\hspace{6pt}}cccccccc}
\toprule
\textbf{Item} & \textbf{SI} & \textbf{ZS-CoT} & \textbf{CoT-SC} & \textbf{P\&S} & \textbf{L2M} & \textbf{BoT} & \textbf{NoDiff} & \textbf{Total} \\
\midrule
\multicolumn{9}{l}{\textit{Structural understanding}} \\
Step comprehension       & 6 & \cellcolor{winnerGreen}21 & 0 & 11 & 6  & 3 & 3  & 50 \\
Dependency comprehension & 3 & 6  & 3 & \cellcolor{winnerGreen}15 & 11 & 7 & 5  & 50 \\
Traceability             & 5 & \cellcolor{winnerGreen}18 & 0 & 10 & 5  & 8 & 4  & 50 \\
\midrule
\multicolumn{9}{l}{\textit{Error localization and detection}} \\
Error detectability      & 4 & \cellcolor{winnerGreen}16 & 2 & 6  & 9  & 5 & 8  & 50 \\
Error localizability     & 4 & 10 & 2 & 5  & \cellcolor{winnerGreen}13 & 6 & 10 & 50 \\
\midrule
\multicolumn{9}{l}{\textit{Trust and reliance}} \\
Reliability              & 3 & 10 & 1 & \cellcolor{winnerGreen}14 & 8  & 7 & 7  & 50 \\
Decision confidence      & 2 & 4  & 2 & \cellcolor{winnerGreen}16 & 15 & 6 & 5  & 50 \\
Critical evaluation      & 2 & 6  & 4 & 10 & \cellcolor{winnerGreen}16 & 5 & 7  & 50 \\
\midrule
\textbf{Total}           & \textbf{29} & \textbf{91} & \textbf{14} & \textbf{87} & \textbf{83} & \textbf{47} & \textbf{49} & \textbf{400} \\
\bottomrule
\end{tabular}
\caption{%
\textbf{Post-survey item-level selection counts across the six reasoning representations} (\textit{SI}: Standard I/O; \textit{ZS-CoT}: Zero-shot CoT; \textit{CoT-SC}: CoT Self-Consistency; \textit{P\&S}: Plan-and-Solve; \textit{L2M}: Least-to-Most; \textit{BoT}: Buffer of Thoughts; 
\textit{NoDiff}: No difference perceived).
The most %
selected option
for each item is highlighted in 
\textcolor{green}{green}. %
}
\label{tab:post_survey_counts}
\end{table*}

\topic{Structure helps participants 
identify reasoning shifts, but not which steps matter.}
Table~\ref{tab:rq1_summary} reports
whether participants perceive the structural 
organization of each reasoning representation.
We distinguish between local cues---%
which steps are key or supportive---%
and a global cue: where the reasoning shifts.
Agreement on local cues is modest 
across all representations 
(key-step $\alpha\!=\!0.11$--$0.37$; 
 support-step $\alpha\!=\!0.05$--$0.48$).
Participants 
do not consistently agree on which steps matter
or how they support one another.
For the global cue, shift recognition is low throughout: 
even Plan-and-Solve and Buffer of Thoughts---%
designed to expose explicit planning structure---%
reach only 60\% and 64\%, so roughly two in five participants fail to identify a reasoning shift 
even when the representation marks it.
Planning-oriented representations 
yield the highest shift recognition, 
suggesting that participants perceive organizational shifts
such as planning-to-execution or template-to-instantiation.
Standard I/O instead yields the strongest support-step agreement ($\alpha\!=\!0.48$). 
Least-to-Most shows the opposite: moderate shift recognition but the weakest support-step agreement 
($\alpha\!=\!0.05$). 
Overall, decomposition makes 
the organization of reasoning more visible, 
but does not necessarily make 
the importance of individual steps 
or their dependencies easier to interpret.

\begin{table}[ht]
\centering
\small
\setlength{\tabcolsep}{7pt}
\renewcommand{\arraystretch}{1.15}
\adjustbox{max width=\linewidth}{
\begin{tabular}{lccccc}
\toprule
\textbf{Representation} 
& \textbf{Pref.} $\uparrow$
& \textbf{FA} $\downarrow$
& \textbf{Loc.} $\uparrow$
& \textbf{Reliance} $\uparrow$
& \textbf{Interp.} $\uparrow$ \\
\midrule \midrule
Standard I/O       & 6.0\%  & 12.5\% & 60.0\% & 3.32 & 3.66 \\
Zero-shot CoT      & 22.0\% & \textbf{7.7\%}  & \textbf{95.5\%} & \textbf{4.04} & \textbf{4.12} \\
CoT-SC             & 0.0\%  & 12.5\% & 73.2\% & 3.37 & 3.61 \\
Plan-and-Solve     & \textbf{26.0\%} & 19.2\% & 57.1\% & 3.88 & 3.62 \\
Least-to-Most      & 22.0\% & 20.8\% & 90.5\% & 3.52 & 3.28 \\
Buffer of Thoughts & 14.0\% & 17.4\% & 52.2\% & 3.54 & 3.08 \\
\bottomrule
\end{tabular}
}
\caption{
\textbf{%
Summarizing our results across reasoning representations.}
\emph{Preference} reports overall preference votes.
\emph{FA} is the false-alarm rate on correct traces, and \emph{Loc.} is localization accuracy on errored traces.
\emph{Reliance} and \emph{Interp.} are mean 5-point Likert ratings.
\textbf{Bold} indicates the most favorable value in each column.
Plan-and-Solve is preferred most yet has one of the highest false-alarm rates, whereas Zero-shot CoT achieves the strongest %
performance across the three dimensions.
}
\label{tab:heatmap_divergence}
\end{table}

\topic{Zero-shot CoT performs best on 
error verification and trust (D2--D3), 
while not being the most preferred representation.}
Table~\ref{tab:heatmap_divergence}
summarizes our results.
On error-injected trials, 
it yields the lowest false-alarm rate 
(7.7\%, incorrectly flagging correct traces as erroneous) 
and the highest localization accuracy 
(95.5\%, correctly identifying the injected step). 
On the 5-point Likert-scale items 
for reliance and interpretability,
it receives the highest mean ratings for reliance (4.04) 
and interpretability (4.12). 
In the post-survey
(see Table~\ref{tab:post_survey_counts}),
it is the most frequently chosen representation for step comprehension (21/50), traceability(18/50), and error detection (16/50), although Plan-and-Solve receives more overall-preference votes (26\% vs.\ 22\%).
The least structured representation, Zero-shot CoT, 
performs best across verification, trust,
and perception simultaneously.

\topic{Representations rated best for verification in the post-survey have the highest false alarm rates.}
Participants rate Least-to-Most (L2M) as the one that best supports %
error detection and localization in the post-survey:
it receives the most vote for
critical evaluation (16/50) and error localizability (13/50) in Table~\ref{tab:post_survey_counts}.
It also ties with Zero-shot CoT for second place in overall preference (22\%; Table~\ref{tab:heatmap_divergence}). 
Yet on error-injected trials, L2M shows the highest false-alarm rate of all six representations (20.8\%), about 2.7 times that of Zero-shot CoT (7.7\%). 
The pattern extends to Plan-and-Solve, the most preferred representation on the post-survey overall, which shows the second-highest false-alarm rate (19.2\%).
The representation participants pick as best for 
catching mistakes is the one on which 
they most often see errors that are not there---%
stated preference diverges sharply from behavioral accuracy.

\section{Conclusion}
\label{sec:conclusion}

This work studies reasoning representations as interfaces for human oversight,
rather than as model-centric indicators of its ability alone. 
Through a controlled human evaluation of six reasoning representations, 
we show that \emph{user preference, verification performance, and trust calibration do not always align}:
users favor more structured planning- and decomposition-based ones, 
while simpler CoT traces better support error detection and localization. 
Preferred formats can also produce miscalibrated judgments,
suggesting that perceived usefulness does not guarantee effective verification.
Our findings challenge the assumption 
that more visible or structured reasoning uniformly improves human evaluation.
As reasoning becomes embedded in user-facing LLM systems,
its representations should be designed and evaluated 
for the human judgments they support---%
understanding, verification, and trust calibration---%
rather than for model-centric criteria such as accuracy or faithfulness.

\section*{Acknowledgments}
\label{sec:ack}

We thank the anonymous reviewers for constructive feedback.
Jaewoo and Sanghyun are partially supported by
the Google Faculty Research Award 2023.
Sungbok is in part supported by Institute of Information \& communications Technology Planning \& Evaluation (IITP) under the Artificial Intelligence Innovation Human Resources Development (IITP-RS-2026-25547954)
grant funded by the Korea government.
The findings and conclusions in this work 
are those of the author(s) and 
do not necessarily represent 
the views of the funding agency.

\section*{Limitations}

Our study evaluates pre-generated LLM outputs
rather than reasoning traces produced during interactive sessions.
This may limit the forms of interaction available to users,
such as asking follow-up questions, 
requesting alternative explanations, 
or revising their judgments through dialogue.
We choose this design, consistent with prior controlled studies~\cite{zhou2025icot},
to enable matched comparisons 
across task, model, problem instance, 
and representation format.
This allows us to isolate representation-level effects
that would be difficult to measure in open-ended interaction.
We leave the extension to interactive settings 
for future work.

Our findings may be influenced by
how reasoning traces and errors are constructed.
Because each representation uses 
a different prompting strategy, differences in 
trace length, wording, level of detail, and structure
may partly reflect prompt design 
rather than the representation itself.
Our controlled error injection provides 
ground truth for error detection and localization, 
yet verification behavior may depend on 
the type and location of the injected error.
This procedure also requires a localizable reasoning trace,
limiting its applicability to formats
such as Standard I/O.
We mitigate these issues 
by holding the task, model, and problem fixed 
and applying the same evaluation protocol 
across representations.
We leave alternative prompt templates 
and broader error taxonomies to future work.

All traces are generated by a single model (GPT-5).
Using a single generator controls for model-level variation,
allowing us to attribute observed differences 
more directly to the reasoning representation.
However, this design does not evaluate 
whether our findings generalize across models.
We leave cross-model replication for future work.

\section*{Ethics Considerations}
\label{sec:ethics}

This study was determined to be exempt 
by the Oregon State University Institutional Review Board
(protocol \#HE-2025-1705).
Participation was voluntary, 
no identifying information was collected, 
and all responses were stored in anonymized form.

\bibliography{bib/thiswork}

\begin{thebibliography}{53}
\providecommand{\natexlab}[1]{#1}

\bibitem[{Bansal et~al.(2021)Bansal, Wu, Zhou, Fok, Nushi, Kamar, Ribeiro, and
  Weld}]{bansal2021does}
Gagan Bansal, Tongshuang Wu, Joyce Zhou, Raymond Fok, Besmira Nushi, Ece Kamar,
  Marco~Tulio Ribeiro, and Daniel Weld. 2021.
\newblock \href {https://doi.org/10.1145/3411764.3445717} {Does the whole
  exceed its parts? the effect of {AI} explanations on complementary team
  performance}.
\newblock In \emph{Proceedings of the 2021 CHI Conference on Human Factors in
  Computing Systems (CHI)}, pages 1--16. Association for Computing Machinery.

\bibitem[{Besta et~al.(2024{\natexlab{a}})Besta, Blach, Kubicek, Gerstenberger,
  Podstawski, Gianinazzi, Gajda, Lehmann, Niewiadomski, Nyczyk, and
  Hoefler}]{besta2024graph}
Maciej Besta, Nils Blach, Ales Kubicek, Robert Gerstenberger, Micha{\l}
  Podstawski, Lukas Gianinazzi, Joanna Gajda, Tomasz Lehmann, Hubert
  Niewiadomski, Piotr Nyczyk, and Torsten Hoefler. 2024{\natexlab{a}}.
\newblock Graph of thoughts: Solving elaborate problems with large language
  models.
\newblock \emph{Proceedings of the AAAI Conference on Artificial Intelligence},
  38(16):17682--17690.

\bibitem[{Besta et~al.(2024{\natexlab{b}})Besta, Memedi, Zhang, Gerstenberger,
  Piao, Blach, Nyczyk, Copik, Kwa{\'s}niewski, M{\"u}ller
  et~al.}]{besta2024demystifying}
Maciej Besta, Florim Memedi, Zhenyu Zhang, Robert Gerstenberger, Guangyuan
  Piao, Nils Blach, Piotr Nyczyk, Marcin Copik, Grzegorz Kwa{\'s}niewski,
  J{\"u}rgen M{\"u}ller, and 1 others. 2024{\natexlab{b}}.
\newblock Demystifying chains, trees, and graphs of thoughts.
\newblock \emph{arXiv preprint arXiv:2401.14295}.

\bibitem[{Bogdan et~al.(2025)Bogdan, Macar, Nanda, and
  Conmy}]{bogdan2025thought}
Paul~C. Bogdan, Uzay Macar, Neel Nanda, and Arthur Conmy. 2025.
\newblock Thought anchors: Which {LLM} reasoning steps matter?
\newblock \emph{arXiv preprint arXiv:2506.19143}.

\bibitem[{Brown et~al.(2020)Brown, Mann, Ryder, Subbiah, Kaplan, Dhariwal,
  Neelakantan, Shyam, Sastry, Askell et~al.}]{brown2020language}
Tom Brown, Benjamin Mann, Nick Ryder, Melanie Subbiah, Jared~D. Kaplan,
  Prafulla Dhariwal, Arvind Neelakantan, Pranav Shyam, Girish Sastry, Amanda
  Askell, and 1 others. 2020.
\newblock Language models are few-shot learners.
\newblock In \emph{Advances in Neural Information Processing Systems},
  volume~33, pages 1877--1901.

\bibitem[{Bu{\c{c}}inca et~al.(2020)Bu{\c{c}}inca, Lin, Gajos, and
  Glassman}]{buccinca20proxy}
Zana Bu{\c{c}}inca, Phoebe Lin, Krzysztof~Z. Gajos, and Elena~L. Glassman.
  2020.
\newblock \href {https://doi.org/10.1145/3377325.3377498} {Proxy tasks and
  subjective measures can be misleading in evaluating explainable {AI}
  systems}.
\newblock In \emph{International Conference on Intelligent User Interfaces},
  pages 454--464. {ACM}.

\bibitem[{Bu{\c{c}}inca et~al.(2021)Bu{\c{c}}inca, Malaya, and
  Gajos}]{bucinca2021trust}
Zana Bu{\c{c}}inca, Maja~Barbara Malaya, and Krzysztof~Z. Gajos. 2021.
\newblock \href {https://doi.org/10.1145/3449287} {To trust or to think:
  Cognitive forcing functions can reduce overreliance on {AI} in {AI}-assisted
  decision-making}.
\newblock \emph{Proceedings of the ACM on Human-Computer Interaction},
  5(CSCW1):1--21.

\bibitem[{Chen et~al.(2023{\natexlab{a}})Chen, Ma, Wang, and
  Cohen}]{chen2023program}
Wenhu Chen, Xueguang Ma, Xinyi Wang, and William~W. Cohen. 2023{\natexlab{a}}.
\newblock Program of thoughts prompting: Disentangling computation from
  reasoning for numerical reasoning tasks.
\newblock \emph{Transactions on Machine Learning Research}.

\bibitem[{Chen et~al.(2023{\natexlab{b}})Chen, Lin, Sch{\"a}rli, and
  Zhou}]{chen2023selfdebug}
Xinyun Chen, Maxwell Lin, Nathanael Sch{\"a}rli, and Denny Zhou.
  2023{\natexlab{b}}.
\newblock Teaching large language models to self-debug.
\newblock \emph{arXiv preprint arXiv:2304.05128}.

\bibitem[{Cobbe et~al.(2021)Cobbe, Kosaraju, Bavarian, Chen, Jun, Kaiser,
  Plappert, Tworek, Hilton, Nakano et~al.}]{cobbe2021gsm8k}
Karl Cobbe, Vineet Kosaraju, Mohammad Bavarian, Mark Chen, Heewoo Jun, Lukasz
  Kaiser, Matthias Plappert, Jerry Tworek, Jacob Hilton, Reiichiro Nakano, and
  1 others. 2021.
\newblock Training verifiers to solve math word problems.
\newblock \emph{arXiv preprint arXiv:2110.14168}.

\bibitem[{Fok and Weld(2024)}]{fok2024verifiability}
Raymond Fok and Daniel~S. Weld. 2024.
\newblock In search of verifiability: Explanations rarely enable complementary
  performance in {AI}-advised decision making.
\newblock \emph{AI Magazine}, 45(3):317--332.

\bibitem[{Golovneva et~al.(2023)Golovneva, Chen, Poff, Corredor, Zettlemoyer,
  Fazel-Zarandi, and Celikyilmaz}]{golovneva2023roscoe}
Olga Golovneva, Moya Chen, Spencer Poff, Martin Corredor, Luke Zettlemoyer,
  Maryam Fazel-Zarandi, and Asli Celikyilmaz. 2023.
\newblock \href {https://openreview.net/forum?id=xYlJRpzZtsY} {{ROSCOE}: A
  suite of metrics for scoring step-by-step reasoning}.
\newblock In \emph{International Conference on Learning Representations
  (ICLR)}.

\bibitem[{Goyal et~al.(2024)Goyal, Ji, Rawat, Menon, Kumar, and
  Nagarajan}]{goyal2024pause}
Sachin Goyal, Ziwei Ji, Ankit~Singh Rawat, Aditya~Krishna Menon, Sanjiv Kumar,
  and Vaishnavh Nagarajan. 2024.
\newblock Think before you speak: Training language models with pause tokens.
\newblock In \emph{International Conference on Learning Representations
  (ICLR)}.

\bibitem[{Hao et~al.(2024)Hao, Sukhbaatar, Su, Li, Hu, Weston, and
  Tian}]{hao2024coconut}
Shibo Hao, Sainbayar Sukhbaatar, DiJia Su, Xian Li, Zhiting Hu, Jason Weston,
  and Yuandong Tian. 2024.
\newblock Training large language models to reason in a continuous latent
  space.
\newblock \emph{arXiv preprint arXiv:2412.06769}.

\bibitem[{Hendrycks et~al.(2021)Hendrycks, Burns, Kadavath, Arora, Basart,
  Tang, Song, and Steinhardt}]{hendrycks21mathdataset}
Dan Hendrycks, Collin Burns, Saurav Kadavath, Akul Arora, Steven Basart, Eric
  Tang, Dawn Song, and Jacob Steinhardt. 2021.
\newblock Measuring mathematical problem solving with the {MATH} dataset.
\newblock In \emph{Proceedings of the Neural Information Processing Systems
  Track on Datasets and Benchmarks}.

\bibitem[{Hoffman et~al.(2023)Hoffman, Mueller, Klein, and
  Litman}]{hoffman2023measures}
Robert~R. Hoffman, Shane~T. Mueller, Gary Klein, and Jordan Litman. 2023.
\newblock Measures for explainable {AI}: Explanation goodness, user
  satisfaction, mental models, curiosity, trust, and human-{AI} performance.
\newblock \emph{Frontiers in Computer Science}, 5:1096257.

\bibitem[{Jian et~al.(2000)Jian, Bisantz, and Drury}]{jian2000foundations}
Jiun-Yin Jian, Ann~M. Bisantz, and Colin~G. Drury. 2000.
\newblock Foundations for an empirically determined scale of trust in automated
  systems.
\newblock \emph{International Journal of Cognitive Ergonomics}, 4(1):53--71.

\bibitem[{Kamoi et~al.(2024)Kamoi, Das, Lou, Ahn, Zhao, Lu, Zhang, Zhang,
  Zhang, Vummanthala et~al.}]{kamoi2024evaluating}
Ryo Kamoi, Sarkar Snigdha~Sarathi Das, Renze Lou, Jihyun~Janice Ahn, Yilun
  Zhao, Xiaoxin Lu, Nan Zhang, Yusen Zhang, Ranran~Haoran Zhang, Sujeeth~Reddy
  Vummanthala, and 1 others. 2024.
\newblock Evaluating {LLM}s at detecting errors in {LLM} responses.
\newblock \emph{arXiv preprint arXiv:2404.03602}.

\bibitem[{Khot et~al.(2023)Khot, Trivedi, Finlayson, Fu, Richardson, Clark, and
  Sabharwal}]{khot2023decomposed}
Tushar Khot, Harsh Trivedi, Matthew Finlayson, Yao Fu, Kyle Richardson, Peter
  Clark, and Ashish Sabharwal. 2023.
\newblock Decomposed prompting: A modular approach for solving complex tasks.
\newblock In \emph{International Conference on Learning Representations
  (ICLR)}.

\bibitem[{Kim et~al.(2024)Kim, Liao, Vorvoreanu, Ballard, and
  Vaughan}]{kim24llmuncertainty}
Sunnie S.~Y. Kim, Q.~Vera Liao, Mihaela Vorvoreanu, Stephanie Ballard, and
  Jennifer~Wortman Vaughan. 2024.
\newblock \href {https://doi.org/10.1145/3630106.3658941} {"i'm not sure,
  but...": Examining the impact of large language models' uncertainty
  expression on user reliance and trust}.
\newblock In \emph{Proceedings of the {ACM} Conference on Fairness,
  Accountability, and Transparency}, pages 822--835. {ACM}.

\bibitem[{Kim et~al.(2025)Kim, Vaughan, Liao, Lombrozo, and
  Russakovsky}]{kim25llmreliance}
Sunnie S.~Y. Kim, Jennifer~Wortman Vaughan, Q.~Vera Liao, Tania Lombrozo, and
  Olga Russakovsky. 2025.
\newblock \href {https://doi.org/10.1145/3706598.3714020} {Fostering
  appropriate reliance on large language models: The role of explanations,
  sources, and inconsistencies}.
\newblock In \emph{Proceedings of the 2025 ACM Conference on Human Factors in
  Computing Systems}, pages 420:1--420:19. {ACM}.

\bibitem[{Kojima et~al.(2022)Kojima, Gu, Reid, Matsuo, and
  Iwasawa}]{kojima2022large}
Takeshi Kojima, Shixiang~Shane Gu, Machel Reid, Yutaka Matsuo, and Yusuke
  Iwasawa. 2022.
\newblock Large language models are zero-shot reasoners.
\newblock In \emph{Advances in Neural Information Processing Systems},
  volume~35, pages 22199--22213.

\bibitem[{Lanchantin et~al.(2023)Lanchantin, Toshniwal, Weston, Szlam, and
  Sukhbaatar}]{lanchantin2023learning}
Jack Lanchantin, Shubham Toshniwal, Jason Weston, Arthur Szlam, and Sainbayar
  Sukhbaatar. 2023.
\newblock Learning to reason and memorize with self-notes.
\newblock In \emph{Advances in Neural Information Processing Systems
  (NeurIPS)}.

\bibitem[{Lanham et~al.(2023)Lanham, Chen, Radhakrishnan, Steiner, and
  et~al.}]{lanham23faithfulness}
Tamera Lanham, Anna Chen, Ansh Radhakrishnan, Benoit Steiner, and et~al. 2023.
\newblock \href {https://doi.org/10.48550/ARXIV.2307.13702} {Measuring
  faithfulness in chain-of-thought reasoning}.
\newblock \emph{CoRR}.

\bibitem[{Lee and Hockenmaier(2025)}]{lee2025evaluating}
Jinu Lee and Julia Hockenmaier. 2025.
\newblock Evaluating step-by-step reasoning traces: A survey.
\newblock In \emph{Findings of the Association for Computational Linguistics:
  EMNLP 2025}, pages 1789--1814, Suzhou, China. Association for Computational
  Linguistics.

\bibitem[{Lightman et~al.(2024)Lightman, Kosaraju, Burda, Edwards, Baker, Lee,
  Leike, Schulman, Sutskever, and Cobbe}]{lightman24stepbystep}
Hunter Lightman, Vineet Kosaraju, Yuri Burda, Harrison Edwards, Bowen Baker,
  Teddy Lee, Jan Leike, John Schulman, Ilya Sutskever, and Karl Cobbe. 2024.
\newblock Let's verify step by step.
\newblock In \emph{The International Conference on Learning Representations}.

\bibitem[{Lyu et~al.(2023)Lyu, Havaldar, Stein, Zhang, Rao, Wong, Apidianaki,
  and Callison-Burch}]{lyu2023faithful}
Qing Lyu, Shreya Havaldar, Adam Stein, Li~Zhang, Delip Rao, Eric Wong, Marianna
  Apidianaki, and Chris Callison-Burch. 2023.
\newblock Faithful chain-of-thought reasoning.
\newblock In \emph{Proceedings of the 13th International Joint Conference on
  Natural Language Processing and the 3rd Conference of the Asia-Pacific
  Chapter of the Association for Computational Linguistics (Volume 1: Long
  Papers)}, pages 305--329.

\bibitem[{Madaan et~al.(2023)Madaan, Tandon, Gupta, Hallinan, Gao, Wiegreffe,
  Alon, Dziri, Prabhumoye, Yang et~al.}]{madaan2023selfrefine}
Aman Madaan, Niket Tandon, Prakhar Gupta, Skyler Hallinan, Luyu Gao, Sarah
  Wiegreffe, Uri Alon, Nouha Dziri, Shrimai Prabhumoye, Yiming Yang, and 1
  others. 2023.
\newblock Self-refine: Iterative refinement with self-feedback.
\newblock In \emph{Advances in Neural Information Processing Systems},
  volume~36.

\bibitem[{Madsen et~al.(2024)Madsen, Chandar, and Reddy}]{madsen2024self}
Andreas Madsen, Sarath Chandar, and Siva Reddy. 2024.
\newblock \href {https://aclanthology.org/2024.findings-acl.19/} {Are
  self-explanations from large language models faithful?}
\newblock In \emph{Findings of the Association for Computational Linguistics:
  ACL 2024}, pages 295--337, Bangkok, Thailand. Association for Computational
  Linguistics.

\bibitem[{Miao et~al.(2024)Miao, Teh, and Rainforth}]{miao2024selfcheck}
Ning Miao, Yee~Whye Teh, and Tom Rainforth. 2024.
\newblock {SelfCheck}: Using {LLM}s to zero-shot check their own step-by-step
  reasoning.
\newblock In \emph{International Conference on Learning Representations
  (ICLR)}.

\bibitem[{Ning et~al.(2024)Ning, Lin, Zhou, Wang, Yang, and Wang}]{ning2023sot}
Xuefei Ning, Zinan Lin, Zixuan Zhou, Zifu Wang, Huazhong Yang, and Yu~Wang.
  2024.
\newblock Skeleton-of-thought: Prompting {LLM}s for efficient parallel
  generation.
\newblock In \emph{International Conference on Learning Representations
  (ICLR)}.

\bibitem[{Nye et~al.(2021)Nye, Andreassen, Gur-Ari, Michalewski, Austin,
  Bieber, Dohan, Lewkowycz, Bosma, Luan et~al.}]{nye2021show}
Maxwell Nye, Anders~Johan Andreassen, Guy Gur-Ari, Henryk Michalewski, Jacob
  Austin, David Bieber, David Dohan, Aitor Lewkowycz, Maarten Bosma, David
  Luan, and 1 others. 2021.
\newblock Show your work: Scratchpads for intermediate computation with
  language models.
\newblock \emph{arXiv preprint arXiv:2112.00114}.

\bibitem[{Pang et~al.(2026)Pang, Feng, Feng, Li, Shi, Tsvetkov, Heer, and
  Reinecke}]{pang26interactivereasoning}
Rock~Yuren Pang, K.~J.~Kevin Feng, Shangbin Feng, Chu Li, Weijia Shi, Yulia
  Tsvetkov, Jeffrey Heer, and Katharina Reinecke. 2026.
\newblock \href {https://doi.org/10.1145/3742413.3789091} {Interactive
  reasoning: Visualizing and controlling chain-of-thought reasoning in large
  language models}.
\newblock In \emph{Proceedings of the International Conference on Intelligent
  User Interfaces}, pages 852--867. {ACM}.

\bibitem[{Prasad et~al.(2023)Prasad, Saha, Zhou, and
  Bansal}]{prasad2023receval}
Archiki Prasad, Swarnadeep Saha, Xiang Zhou, and Mohit Bansal. 2023.
\newblock \href {https://aclanthology.org/2023.emnlp-main.622/} {{ReCEval}:
  Evaluating reasoning chains via correctness and informativeness}.
\newblock In \emph{Proceedings of the 2023 Conference on Empirical Methods in
  Natural Language Processing (EMNLP)}, pages 10066--10086, Singapore.
  Association for Computational Linguistics.

\bibitem[{Sel et~al.(2024)Sel, Al-Tawaha, Khattar, Jia, and
  Jin}]{sel2024algorithm}
Bilgehan Sel, Ahmad Al-Tawaha, Vanshaj Khattar, Ruoxi Jia, and Ming Jin. 2024.
\newblock Algorithm of thoughts: Enhancing exploration of ideas in large
  language models.
\newblock In \emph{International Conference on Machine Learning (ICML)}.

\bibitem[{Shinn et~al.(2023)Shinn, Cassano, Gopinath, Narasimhan, and
  Yao}]{shinn2023reflexion}
Noah Shinn, Federico Cassano, Ashwin Gopinath, Karthik Narasimhan, and Shunyu
  Yao. 2023.
\newblock Reflexion: Language agents with verbal reinforcement learning.
\newblock In \emph{Advances in Neural Information Processing Systems},
  volume~36.

\bibitem[{Si et~al.(2024)Si, Goyal, Wu, Zhao, Feng, III, and
  Boyd{-}Graber}]{si24llmstruthfulness}
Chenglei Si, Navita Goyal, Tongshuang Wu, Chen Zhao, Shi Feng, Hal~Daum{\'{e}}
  III, and Jordan~L. Boyd{-}Graber. 2024.
\newblock \href {https://doi.org/10.18653/V1/2024.NAACL-LONG.81} {Large
  language models help humans verify truthfulness - except when they are
  convincingly wrong}.
\newblock In \emph{Proceedings of the 2024 Conference of the North American
  Chapter of the Association for Computational Linguistics: Human Language
  Technologies}, pages 1459--1474.

\bibitem[{Sprague et~al.(2025)Sprague, Yin, Rodriguez, Jiang, Wadhwa, Singhal,
  Zhao, Ye, Mahowald, and Durrett}]{sprague2024cot}
Zayne Sprague, Fangcong Yin, Juan~Diego Rodriguez, Dongwei Jiang, Manya Wadhwa,
  Prasann Singhal, Xinyu Zhao, Xi~Ye, Kyle Mahowald, and Greg Durrett. 2025.
\newblock To {CoT} or not to {CoT}? chain-of-thought helps mainly on math and
  symbolic reasoning.
\newblock In \emph{International Conference on Learning Representations
  (ICLR)}.

\bibitem[{Sun et~al.(2026)Sun, Wei, Bosch, Echizen, Sugawara, and
  Ali}]{sun26seeingreasoning}
Xin Sun, Shu Wei, Jos~A. Bosch, Isao Echizen, Saku Sugawara, and Abdallah~El
  Ali. 2026.
\newblock \href {https://doi.org/10.1145/3772363.3798613} {Seeing the
  reasoning: How {LLM} rationales influence user trust and decision-making in
  factual verification tasks}.
\newblock In \emph{Proceedings of the Extended Abstracts of the 2026 ACM
  Conference on Human Factors in Computing Systems}, pages 585:1--585:7. {ACM}.

\bibitem[{Suzgun et~al.(2023)Suzgun, Scales, Sch{\"a}rli, Gehrmann, Tay, Chung,
  Chowdhery, Le, Chi, Zhou, and Wei}]{suzgun2023bbh}
Mirac Suzgun, Nathan Scales, Nathanael Sch{\"a}rli, Sebastian Gehrmann, Yi~Tay,
  Hyung~Won Chung, Aakanksha Chowdhery, Quoc~V. Le, Ed~H. Chi, Denny Zhou, and
  Jason Wei. 2023.
\newblock Challenging {BIG}-bench tasks and whether chain-of-thought can solve
  them.
\newblock In \emph{Findings of the Association for Computational Linguistics:
  ACL 2023}, pages 13003--13051.

\bibitem[{Turpin et~al.(2023)Turpin, Michael, Perez, and
  Bowman}]{turpin2023language}
Miles Turpin, Julian Michael, Ethan Perez, and Samuel~R. Bowman. 2023.
\newblock Language models don't always say what they think: Unfaithful
  explanations in chain-of-thought prompting.
\newblock In \emph{Advances in Neural Information Processing Systems},
  volume~36.

\bibitem[{Vasconcelos et~al.(2023)Vasconcelos, J{\"o}rke, Grunde-McLaughlin,
  Gerstenberg, Bernstein, and Krishna}]{vasconcelos2023explanations}
Helena Vasconcelos, Matthew J{\"o}rke, Madeleine Grunde-McLaughlin, Tobias
  Gerstenberg, Michael~S. Bernstein, and Ranjay Krishna. 2023.
\newblock Explanations can reduce overreliance on {AI} systems during
  decision-making.
\newblock \emph{Proceedings of the ACM on Human-Computer Interaction},
  7(CSCW1).

\bibitem[{Wang et~al.(2023{\natexlab{a}})Wang, Xu, Lan, Hu, Lan, Lee, and
  Lim}]{wang2023plansolve}
Lei Wang, Wanyu Xu, Yihuai Lan, Zhiqiang Hu, Yunshi Lan, Roy Ka-Wei Lee, and
  Ee-Peng Lim. 2023{\natexlab{a}}.
\newblock Plan-and-solve prompting: Improving zero-shot chain-of-thought
  reasoning by large language models.
\newblock In \emph{Proceedings of the 61st Annual Meeting of the Association
  for Computational Linguistics (Volume 1: Long Papers)}, pages 2609--2634.

\bibitem[{Wang et~al.(2023{\natexlab{b}})Wang, Wei, Schuurmans, Le, Chi,
  Narang, Chowdhery, and Zhou}]{wang2023selfconsistency}
Xuezhi Wang, Jason Wei, Dale Schuurmans, Quoc~V Le, Ed~H. Chi, Sharan Narang,
  Aakanksha Chowdhery, and Denny Zhou. 2023{\natexlab{b}}.
\newblock Self-consistency improves chain of thought reasoning in language
  models.
\newblock In \emph{International Conference on Learning Representations
  (ICLR)}.

\bibitem[{Wei et~al.(2022)Wei, Wang, Schuurmans, Bosma, Ichter, Xia, Chi, Le,
  and Zhou}]{wei2022}
Jason Wei, Xuezhi Wang, Dale Schuurmans, Maarten Bosma, Brian Ichter, Fei Xia,
  Ed~H. Chi, Quoc~V. Le, and Denny Zhou. 2022.
\newblock Chain-of-thought prompting elicits reasoning in large language
  models.
\newblock In \emph{Advances in Neural Information Processing Systems},
  volume~35, pages 24824--24837.

\bibitem[{Yang et~al.(2024)Yang, Yu, Zhang, Cao, Xu, Zhang, Gonzalez, and
  Cui}]{yang2024buffer}
Ling Yang, Zhaochen Yu, Tianjun Zhang, Shiyi Cao, Minkai Xu, Wentao Zhang,
  Joseph~E. Gonzalez, and Bin Cui. 2024.
\newblock Buffer of thoughts: Thought-augmented reasoning with large language
  models.
\newblock In \emph{Advances in Neural Information Processing Systems
  (NeurIPS)}.

\bibitem[{Yang et~al.(2018)Yang, Qi, Zhang, Bengio, Cohen, Salakhutdinov, and
  Manning}]{yang2018hotpotqa}
Zhilin Yang, Peng Qi, Saizheng Zhang, Yoshua Bengio, William~W. Cohen, Ruslan
  Salakhutdinov, and Christopher~D. Manning. 2018.
\newblock {HotpotQA}: A dataset for diverse, explainable multi-hop question
  answering.
\newblock In \emph{Proceedings of the 2018 Conference on Empirical Methods in
  Natural Language Processing (EMNLP)}, pages 2369--2380.

\bibitem[{Yao et~al.(2023{\natexlab{a}})Yao, Yu, Zhao, Shafran, Griffiths, Cao,
  and Narasimhan}]{yao2023tot}
Shunyu Yao, Dian Yu, Jeffrey Zhao, Izhak Shafran, Tom Griffiths, Yuan Cao, and
  Karthik Narasimhan. 2023{\natexlab{a}}.
\newblock Tree of thoughts: Deliberate problem solving with large language
  models.
\newblock In \emph{Advances in Neural Information Processing Systems},
  volume~36.

\bibitem[{Yao et~al.(2023{\natexlab{b}})Yao, Zhao, Yu, Du, Shafran, Narasimhan,
  and Cao}]{yao2023react}
Shunyu Yao, Jeffrey Zhao, Dian Yu, Nan Du, Izhak Shafran, Karthik~R.
  Narasimhan, and Yuan Cao. 2023{\natexlab{b}}.
\newblock {ReAct}: Synergizing reasoning and acting in language models.
\newblock In \emph{International Conference on Learning Representations
  (ICLR)}.

\bibitem[{Zhang et~al.(2023)Zhang, Yang, Yuan, and Yao}]{zhang2023cumulative}
Yifan Zhang, Jingqin Yang, Yang Yuan, and Andrew Chi-Chih Yao. 2023.
\newblock Cumulative reasoning with large language models.
\newblock \emph{arXiv preprint arXiv:2308.04371}.

\bibitem[{Zhou et~al.(2023)Zhou, Sch{\"a}rli, Hou, Wei, Scales, Wang,
  Schuurmans, Cui, Bousquet, Le, and Chi}]{zhou2023leasttomost}
Denny Zhou, Nathanael Sch{\"a}rli, Le~Hou, Jason Wei, Nathan Scales, Xuezhi
  Wang, Dale Schuurmans, Claire Cui, Olivier Bousquet, Quoc Le, and Ed~Chi.
  2023.
\newblock Least-to-most prompting enables complex reasoning in large language
  models.
\newblock In \emph{International Conference on Learning Representations
  (ICLR)}.

\bibitem[{Zhou et~al.(2024)Zhou, Pujara, Ren, Chen, Cheng, Le, Chi, Zhou,
  Mishra, and Zheng}]{zhou2024selfdiscover}
Pei Zhou, Jay Pujara, Xiang Ren, Xinyun Chen, Heng-Tze Cheng, Quoc~V. Le, Ed~H.
  Chi, Denny Zhou, Swaroop Mishra, and Huaixiu~Steven Zheng. 2024.
\newblock Self-discover: Large language models self-compose reasoning
  structures.
\newblock In \emph{Advances in Neural Information Processing Systems
  (NeurIPS)}.

\bibitem[{Zhou et~al.(2025)Zhou, Nguyen, Kharya, Nguyen, and
  Agarwal}]{zhou2025icot}
Runtao Zhou, Giang Nguyen, Nikita Kharya, Anh~Totti Nguyen, and Chirag Agarwal.
  2025.
\newblock Improving human verification of {LLM} reasoning through interactive
  explanation interfaces.
\newblock \emph{arXiv preprint arXiv:2510.22922}.

\end{thebibliography}

\appendix
\makeatletter
\setlength{\@fptop}{0pt}
\setlength{\@dblfptop}{0pt}
\makeatother
\clearpage

\newcommand{\prompthead}[1]{%
  \par\vspace{0.3\baselineskip}\noindent
  {\small\bfseries #1}\par\nopagebreak\vspace{0.2\baselineskip}}
\newcommand{\code}[1]{{\small\ttfamily #1}}
\newenvironment{prompt}%
  {\par\scriptsize\topsep=0pt\partopsep=0pt\verbatim}%
  {\endverbatim\vspace{0.3\baselineskip}}

\section{Detailed Experimental Setup}
\label{app:setup}

Here we describe the generation setup used 
across all six representations (\S\ref{app:setup:generation}), 
the prompt templates (\S\ref{app:prompts}), 
the survey instrument shown to participants (\S\ref{app:survey}), 
and recruitment and consent (\S\ref{app:irb}).

\subsection{Model and Generation Setup}
\label{app:setup:generation}

All traces are generated with OpenAI's GPT-5
(\code{gpt-5-2025-08-07}) through the Chat Completions endpoint,
with temperature $1.0$, %
top-$p$ $1.0$, and at most 8{,}192 completion tokens.
Self-Consistency CoT samples $n=5$ reasoning paths per problem and
aggregates by majority vote; all other representations use a single
sample.
Buffer of Thoughts uses \code{text-embedding-3-large} for template
retrieval (\S\ref{app:prompts}).
We retain only problems for which all six representations produce
the correct final answer: of 140 candidate problems, 27 (19.3\%)
satisfy this filter, 9 per benchmark (Table~\ref{tab:dataset_stats}).

\topic{Modifications from original formulations.}
We applied the following minor adaptations to the original prompt
formulations, in service of consistent answer extraction and a
unified evaluation pipeline:

\begin{itemize}[
    itemsep=0.1em,
    topsep=0.1em,
    leftmargin=1.2em
]
\item \textbf{Self-Consistency CoT:} GPT-5's Chat Completions endpoint
  constrains temperature to $1.0$ and does not accept the lower
  temperatures (e.g., $T=0.7$) used in the original
  work~\citep{wang2023selfconsistency}. We therefore sample at the
  only temperature the endpoint admits and aggregate by majority vote
  over $n=5$ samples.

\item \textbf{Plan-and-Solve:} We add benchmark-specific answer
  extraction triggers (``Therefore, the answer (arabic numerals)
  is''~/~``(a short phrase) is'') in the second stage, in place of
  the single math-oriented trigger from the original paper.

\item \textbf{Buffer of Thoughts:} We add an explicit
  ``Final Answer:'' output format to the buffer-instantiation
  prompt to support deterministic answer extraction across
  benchmarks. The meta-buffer, retrieval mode, and dynamic-update
  procedure are otherwise unchanged from the official
  implementation; we use GPT-5 %
  as the LLM backbone 
  and \code{text-embedding-3-large} %
  for retrieval.

\item \textbf{Least-to-Most:} We retain at most six sub-questions per
  problem and concatenate prior (sub-question, answer) pairs as
  \code{Q:/A:} blocks in the second pass, prepending a short
  \code{(We already know: \ldots)} hint before each sub-question's
  answer slot.
\end{itemize}

\subsection{Prompt Templates}
\label{app:prompts}

Here we present the prompt templates used for the six reasoning
representations chosen in our study.

\topic{Standard I/O.}
Direct question-to-answer template 
that does not trigger any reasoning~\cite{brown2020language}.

\prompthead{GSM8K.}
\begin{prompt}
Q: {question}
A:
\end{prompt}

\prompthead{HotPotQA.}
\begin{prompt}
Context: {context}
Q: {question}
A:
\end{prompt}

\prompthead{BBH.}
\begin{prompt}
{task_description}
Q: {question}
A:
\end{prompt}

\topic{Zero-shot Chain-of-Thought.}
Two-stage template: (1) reasoning trigger, (2) answer extraction by
re-feeding the chain~\cite{kojima2022large,wei2022}.

\prompthead{Stage 1 --- Reasoning (GSM8K, BBH).}
\begin{prompt}
Q: {question}
A: Let's think step by step.
\end{prompt}

\prompthead{Stage 2 --- Answer extraction (GSM8K, BBH).}
\begin{prompt}
Q: {question}
A: Let's think step by step.
{reasoning}

Therefore, the answer is
\end{prompt}

\prompthead{Stage 1 --- Reasoning (HotPotQA).}
\begin{prompt}
Context:
{context}

Q: {question}
A: Let's think step by step.
\end{prompt}

\prompthead{Stage 2 --- Answer extraction (HotPotQA).}
\begin{prompt}
Context:
{context}

Q: {question}
A: Let's think step by step.
{reasoning}

Therefore, the answer is
\end{prompt}

\topic{Self-Consistency CoT.}
8-shot CoT exemplars from \citet{wei2022}, identical to those used by
\citet{wang2023selfconsistency}. The prompt template is shared with
few-shot CoT; only the aggregation rule (majority vote over $n=5$
samples) differs.

\prompthead{GSM8K (8-shot, verbatim from \citealp{wei2022}).}
\begin{prompt}
Q: There are 15 trees in the grove. Grove workers will plant
trees in the grove today. After they are done, there will be
21 trees. How many trees did the grove workers plant today?
A: There are 15 trees originally. Then there were 21 trees
after some more were planted. So there must have been
21 - 15 = 6. The answer is 6.

Q: If there are 3 cars in the parking lot and 2 more cars
arrive, how many cars are in the parking lot?
A: There are originally 3 cars. 2 more cars arrive.
3 + 2 = 5. The answer is 5.

Q: Leah had 32 chocolates and her sister had 42. If they ate
35, how many pieces do they have left in total?
A: Originally, Leah had 32 chocolates. Her sister had 42.
So in total they had 32 + 42 = 74. After eating 35,
they had 74 - 35 = 39. The answer is 39.

Q: Jason had 20 lollipops. He gave Denny some lollipops.
Now Jason has 12 lollipops. How many lollipops did Jason
give to Denny?
A: Jason started with 20 lollipops. Then he had 12 after
giving some to Denny. So he gave Denny 20 - 12 = 8.
The answer is 8.

Q: Shawn has five toys. For Christmas, he got two toys each
from his mom and dad. How many toys does he have now?
A: Shawn started with 5 toys. If he got 2 toys each from his
mom and dad, then that is 4 more toys. 5 + 4 = 9.
The answer is 9.

Q: There were nine computers in the server room. Five more
computers were installed each day, from monday to thursday.
How many computers are now in the server room?
A: There were originally 9 computers. For each of 4 days,
5 more computers were added. So 4 * 5 = 20 computers were
added. 9 + 20 = 29. The answer is 29.

Q: Michael had 58 golf balls. On tuesday, he lost 23 golf
balls. On wednesday, he lost 2 more. How many golf balls did
he have at the end of wednesday?
A: Michael started with 58 golf balls. After losing 23 on
tuesday, he had 58 - 23 = 35. After losing 2 more, he had
35 - 2 = 33. The answer is 33.

Q: Olivia has $23. She bought five bagels for $3 each.
How much money does she have left?
A: Olivia had 23 dollars. 5 bagels for 3 dollars each will
be 5 x 3 = 15 dollars. So she has 23 - 15 = 8 dollars left.
The answer is 8.

Q: {question}
A:
\end{prompt}

\prompthead{HotPotQA (8-shot, multi-hop QA).}
\begin{prompt}
Q: Were Scott Derrickson and Ed Wood of the same
nationality?
A: Scott Derrickson is an American director. Ed Wood was an
American director. So they are of the same nationality.
The answer is: Yes.

Q: What government position was held by the woman who
portrayed Corliss Archer in the film Kiss and Tell?
A: Shirley Temple portrayed Corliss Archer in Kiss and Tell.
Shirley Temple was the US Ambassador to Ghana and to
Czechoslovakia. The answer is: Ambassador.

Q: Are director of film Junglee and director of film Baghban
both from the same country?
A: Junglee was directed by Subodh Mukerji. Baghban was
directed by Ravi Chopra. Both are from India.
The answer is: Yes.

Q: The Oberoi family is part of a hotel company that has a
head office in what city?
A: The Oberoi family is part of The Oberoi Group. The Oberoi
Group has its head office in Delhi, India.
The answer is: Delhi.

Q: What nationality was James Henry Miller's wife?
A: James Henry Miller was the American journalist and social
activist known as Henry Miller. His wife was Lepska,
a Polish-born American. The answer is: Polish-American.

Q: Which magazine was started first, Arthur's Magazine or
First for Women?
A: Arthur's Magazine was started in 1844. First for Women
was started in 1989. So Arthur's Magazine was started first.
The answer is: Arthur's Magazine.

Q: Were Pavel Urysohn and Leonid Levin born in the same
country?
A: Pavel Urysohn was born in Odessa, Russian Empire.
Leonid Levin was born in Dnepropetrovsk, Soviet Union.
Both were born in what was the Russian Empire / Soviet Union.
The answer is: Yes.

Q: Are both Canggu and Seminyak located in Bali?
A: Canggu is a village in Bali. Seminyak is a district in
Bali. Both are located in Bali. The answer is: Yes.

Context:
{context}

Q: {question}
A:
\end{prompt}

\smallskip\noindent\textit{BBH (3-shot, per-subtask exemplars).}
The BBH template instantiates a per-subtask 3-shot block:
\begin{prompt}
{task_description}

{few_shot_examples}
Q: {question}
A:
\end{prompt}
\begin{sloppypar}
Per-subtask 3-shot exemplars are stored in \code{BBH\_FEW\_SHOTS}
for the four subtasks used in our study:
\code{multistep\_arithmetic\_two}, \code{disambiguation\_qa},
\code{date\_understanding}, and
\code{logical\_deduction\_five\_objects}. For example, the
\code{multistep\_arithmetic\_two} block is:
\end{sloppypar}
\begin{prompt}
Q: ((-5 + 9 * -4 - 0) * (4 + -7 + 0 * -5)) =
A: Let's compute step by step. 9 * -4 = -36.
-5 + -36 - 0 = -41. -7 + 0 * -5 = -7. 4 + -7 = -3.
-41 * -3 = 123. The answer is 123.

Q: ((-9 * 7 + -5 + 0) * (-3 + -2 * -4 - -8)) =
A: Let's compute step by step. -9 * 7 = -63.
-63 + -5 + 0 = -68. -2 * -4 = 8.
-3 + 8 - -8 = -3 + 8 + 8 = 13. -68 * 13 = -884.
The answer is -884.

Q: ((3 + 7 * 9 * -3) * (8 + -6 - 2 * -1)) =
A: Let's compute step by step. 7 * 9 = 63.
63 * -3 = -189. 3 + -189 = -186. 2 * -1 = -2.
8 + -6 - -2 = 8 + -6 + 2 = 4. -186 * 4 = -744.
The answer is -744.
\end{prompt}

\topic{Plan-and-Solve (PS+).}
Zero-shot. Two-stage: PS+ trigger then answer extraction. Trigger
taken verbatim from \citet{wang2023plansolve} Table~5 (No.~6).

\prompthead{PS+ trigger (math: GSM8K, BBH).}
\begin{prompt}
Let's first understand the problem, extract relevant
variables and their corresponding numerals, and make a plan.
Then, let's carry out the plan, calculate intermediate
variables (pay attention to correct numerical calculation
and commonsense), solve the problem step by step,
and show the answer.
\end{prompt}

\prompthead{PS+ trigger (QA: HotPotQA).}
\begin{prompt}
Let's first prepare relevant information and make a plan.
Then, let's answer the question step by step (pay attention
to commonsense and logical coherence).
\end{prompt}

\prompthead{Stage 1 --- Reasoning prompt (GSM8K).}
\begin{prompt}
Q: {question}
A: {trigger}
\end{prompt}

\prompthead{Stage 1 --- Reasoning prompt (HotPotQA).}
\begin{prompt}
Context:
{context}

Q: {question}
A: {trigger}
\end{prompt}

\prompthead{Stage 1 --- Reasoning prompt (BBH).}
\begin{prompt}
{task_description}

Q: {question}
A: {trigger}
\end{prompt}

\prompthead{Stage 2 --- Answer extraction (GSM8K).}
\begin{prompt}
Q: {question}
A: {trigger}
{reasoning}

Therefore, the answer (arabic numerals) is
\end{prompt}

\prompthead{Stage 2 --- Answer extraction (HotPotQA).}
\begin{prompt}
Context:
{context}

Q: {question}
A: {trigger}
{reasoning}

Therefore, the answer (a short phrase) is
\end{prompt}

\prompthead{Stage 2 --- Answer extraction (BBH).}
\begin{prompt}
{task_description}

Q: {question}
A: {trigger}
{reasoning}

Therefore, the answer is
\end{prompt}

\topic{Least-to-Most.}
Two-pass template: (1) decomposition, (2) sequential solving with
accumulated prior answers. At most six sub-questions are retained~\cite{zhou2023leasttomost}.

\prompthead{Pass 1 --- Decomposition (GSM8K).}
\begin{prompt}
Q: Elsa has 5 apples. Anna has 2 more apples than Elsa.
How many apples do they have together?
A: To solve "How many apples do they have together?",
I need to first answer:
1. How many apples does Anna have?
2. How many apples do they have together?

Q: {question}
A: To solve this, I need to first answer:
\end{prompt}

\prompthead{Pass 1 --- Decomposition (BBH).}
\begin{prompt}
{task_description}

Q: {question}

Here is an example of how to decompose a similar problem:
Q: {example_question}
A: To solve this, I need to first answer:
{example_subquestions}

Now decompose the above question:
A: To solve this, I need to first answer:
\end{prompt}
\begin{sloppypar}
Example pairs per-subtask are stored in
\code{BBH\_DECOMPOSE\_EXAMPLES}; e.g., for
\code{multistep\_arithmetic\_two}, \code{example\_question} =
\code{\textquotedbl{}((2 + 3) * 4) - 1 =\textquotedbl{}} with three example sub-questions.
\end{sloppypar}

\prompthead{Pass 2 --- Sequential solving (GSM8K).}
\begin{prompt}
Q: Elsa has 5 apples. Anna has 2 more apples than Elsa.
How many apples do they have together?

Q: How many apples does Anna have?
A: Anna has 2 more apples than Elsa. Elsa has 5 apples.
So Anna has 2 + 5 = 7 apples. The answer is: 7.

Q: How many apples do they have together?
(We already know: Anna has 7 apples.)
A: Elsa has 5 apples. Anna has 7 apples. 5 + 7 = 12.
The answer is: 12.

Q: {question}

{solved_context}Q: {sub_question}
{prior_knowledge}A:
\end{prompt}

\prompthead{Pass 2 --- Sequential solving (HotPotQA).}
\begin{prompt}
Context:
{context}

Q: {question}

{solved_context}Q: {sub_question}
{prior_knowledge}A:
\end{prompt}

\prompthead{Pass 2 --- Sequential solving (BBH).}
\begin{prompt}
{task_description}

Q: {question}

{solved_context}Q: {sub_question}
{prior_knowledge}A:
\end{prompt}
\begin{sloppypar}
Here, \code{\{solved\_context\}} concatenates prior (sub-question,
answer) pairs as \code{Q:}/\code{A:} blocks, and
\code{\{prior\_knowledge\}} prepends a single line of the form
\code{(We already know: ...)} listing the answers obtained so far,
before the current sub-question's answer slot.
\end{sloppypar}

\topic{Buffer of Thoughts.}
Three-step pipeline reproducing the official implementation of
\citet{yang2024buffer}, with GPT-5 as the LLM backbone and
\code{text-embedding-3-large} for retrieval.

\prompthead{Step 1 --- Problem Distillation (system prompt).}
\begin{prompt}
As a highly professional and intelligent expert in
information distillation, you excel at extracting essential
information to solve problems from user input queries.
You adeptly transform this extracted information into a
suitable format based on the respective type of the issue.
If the problem can be generalized to a higher level to solve
multiple issues, further analysis and explanation will be
provided upon your next response.
Please categorize and extract the crucial information
required to solve the problem from the user's input query.
Combining these two elements will generate distilled
information. Subsequently, deliver this distilled
information, based on the problem type, to your downstream
meta planner. The problem type should belong to one of the
six categories mentioned above, and the distilled
information should include:

1. Values and information of key variables extracted from
   user input, which will be handed over to the respective
   expert for task resolution, ensuring all essential
   information required to solve the problem is provided.
2. The objective of the problem and corresponding
   constraints.
3. Extend the problem based on 1 and 2, propose a meta
   problem that can address the user query and handle more
   input and output variations. Incorporate the real-world
   scenario of the extended problem along with the types of
   key variables and information constraints from the
   original problem to restrict the key variables in the
   extended problem. After that, use the user query input
   key information as input to solve the problem as an
   example.
4. Try to transform the problem into a python algorithm
   problem, and provide the input parameters.
5. Your task is to distill the problem, you shouldn't give
   the final result or possible solution in your respond.

Please distill the information following the format below
and cease response after the output of the distilled
information.

Meta distiller Respond:

Distilled Information:

1. Key information:

2. Restriction: (It should be noted that the answer should
   strictly follow the real-world rule such as in arithmatic
   equation, the Priority of operator, the need of
   parentheses etc. So according to the distilled
   information, emphasize the real-world rules that need to
   be followed within the problem.)

3. Distilled task:

4. Python transformation:
   (Optional, skip when Python tag is Not for Python) Input
   parameters: (The names of each variable should be clear
   and not confusing, and correspond to the entity names in
   the problem)
     variable1_name = x
     variable2_name = y
     .....
     variableN_name = z

5. Answer form: (Optional, skip when there is no specific
   answer form)

  **Note: The generation ends here. Do not show this
  message in your answer !**
\end{prompt}

\prompthead{Step 2 --- Buffer Instantiation (prefix prepended to the
distilled information).}
\begin{prompt}
You are an expert in problem analysis and can apply previous
problem-solving approaches to new issues. The user will
provide a specific task description and a meta buffer that
holds multiple thought templates that will help to solve
the problem. Your goal is to first extract most relevant
thought template from meta buffer, analyze the user's task
and generate a specific solution based on the thought
template. Give a final answer that is easy to extract from
the text.

IMPORTANT: After your reasoning, conclude your response with
EXACTLY one line in this format:
Final Answer: <your answer>

Where <your answer> is:
- For numerical questions: just the number
  (e.g., "Final Answer: 260")
- For yes/no questions: just yes or no
  (e.g., "Final Answer: yes")
- For multiple-choice questions: the option letter in
  parentheses (e.g., "Final Answer: (E)")
- For short-answer questions: the shortest possible answer
  phrase (e.g., "Final Answer: Albert Einstein")

Do not add any text after the "Final Answer:" line.
\end{prompt}
This text is sent through \code{LightRAG} in \code{hybrid} mode
(graph + vector retrieval over the meta-buffer), which jointly
retrieves the most relevant thought template(s) and generates the
solution in a single call.

\prompthead{Step 3a --- Thought Distillation (system prompt; generates
a new template from the (problem, solution) pair).}
\begin{prompt}
You are an expert in problem analysis and generalization.
Your task is to follow the format of thought template below
and distill a high-level thought template to solve similar
problems:
Example thought template:
### Problem Type 20: Solution Concentration Problem

**Definition**: This type of problem involves the
relationship between a solvent (water or another liquid),
solute, solution, and concentration.

**Quantitative Relationships**:
- Solution = Solvent + Solute
- Concentration = Solute / Solution * 100%

**Solution Strategy**: Use the formulas and their variations
to analyze and calculate the problem.

**Example**: There is 50 grams of a 16%
How much water needs to be added to dilute it to a 10%
sugar solution?

**Solution**:
Using the formula:
50 * 16%

It should be noted that you should only return the thought
template without any extra output.
\end{prompt}

\prompthead{Step 3b --- Dynamic Update similarity judge (prefix
prepended to the newly distilled template).}
\begin{prompt}
Find most relevant thought template in the MetaBuffer
according to the given thought template, and Determine
whether there is a fundamental difference in the
problem-solving approach between this and the most similar
thought template in MetaBuffer. If there is, output "True."
If there is no fundamental difference, or if the two
thought templates are highly similar, output "False."
\end{prompt}
On \code{True}, the new template is inserted into the meta-buffer
via \code{LightRAG.insert}; on \code{False}, it is discarded.

\smallskip\noindent\textit{Meta-buffer contents and retrieval.}
The meta-buffer is initialized from the official \code{math.txt}
shipped with the official Buffer of Thoughts implementation, which
contains \textbf{21 thought templates} (\textit{Problem Type 1:
Normalization Problem} through \textit{Problem Type 21: Equation
Problem}). All 21 templates are indexed by \code{LightRAG} (graph
+ vector store) using \code{text-embedding-3-large} (dim~3072).
Retrieval at inference time uses
\code{QueryParam(mode="hybrid")}, which combines graph-based
entity/relation traversal with dense-vector \textit{k}-NN retrieval
and feeds the retrieved snippets directly into the
buffer-instantiation LLM call. There is no separate top-\textit{k}
cutoff; the relevant templates are selected by the hybrid retriever
within \code{LightRAG}.

\subsection{Survey Instrument}
\label{app:survey}

This section presents the full survey instrument 
shown to participants.
Wording is held constant 
across the six representation conditions.

\topic{Pre-Survey.}
Six items administered after consent; they collected background
covariates only.
\begin{enumerate}[
    noitemsep,
    topsep=0.1em,
    leftmargin=1.2em
]
\item \textbf{Internet Experience.}
``How long have you been using the Internet regularly?''
\begin{itemize}
  \item Less than 2 years
  \item 2--5 years
  \item 5--10 years
  \item More than 10 years
\end{itemize}

\item \textbf{Technical Reading Comfort.}
``How comfortable are you reading technical or analytical material
(e.g., documentation, research articles, or technical explanations)?''
\begin{itemize}
  \item Very uncomfortable
  \item Somewhat uncomfortable
  \item Neutral
  \item Comfortable
  \item Very comfortable
\end{itemize}

\item \textbf{Computer and Programming Familiarity.}
``How would you describe your familiarity with computers and
programming?''
\begin{itemize}
  \item I mainly use computers for everyday tasks (browsing, email,
    documents)
  \item I have some exposure to programming (e.g., classes and
    tutorials)
  \item I occasionally write code for personal projects or work
  \item Programming is a regular part of my academic or professional
    activities
\end{itemize}

\item \textbf{AI Usage Frequency.}
``How often do you use AI-powered tools or assistants (e.g.,
chatbots, writing assistants, recommendation tools)?''
\begin{itemize}
  \item Never
  \item Rarely
  \item Occasionally
  \item Regularly
  \item Very frequently
\end{itemize}

\item \textbf{AI Familiarity.}
``Which of the following best describes your current understanding
of how AI chatbots generate responses?''
\begin{itemize}
  \item I am not familiar with how they work
  \item I have heard about them, but do not know how they work
  \item I have a basic idea of how they generate responses
  \item I understand the general principles behind how they generate
    responses
  \item I have a strong understanding of how these systems generate
    responses
\end{itemize}

\item \textbf{Experience in Evaluating AI Responses.}
``How confident do you usually feel when evaluating whether an
AI-generated response is correct?''

\begin{itemize}
  \item Not confident at all
  \item Slightly confident
  \item Moderately confident
  \item Very confident
  \item Extremely confident
\end{itemize}
\end{enumerate}

\topic{Per-Trace Items.}
Asked once for every representation of each problem; the D1, D2, and
D3 items are asked on the D1, D2, and D3 problems, respectively.
\smallskip\noindent\textit{D1 --- Structural understanding (3 items).}

\begin{enumerate}[
    noitemsep,
    topsep=0.1em,
    leftmargin=1.2em
]
\item \textbf{Reasoning Shift.}
  ``Which step shows that the reasoning changes to a different
  interpretation or approach?''
  Options: enumerated reasoning steps (Step~1, Step~2, \ldots),
  plus ``None'' and ``Not sure.''

\item \textbf{Key Step.}
  ``Which step most directly leads to (or is most important to
  generate) the final answer?''
  Options: enumerated reasoning steps (Step~1, Step~2, \ldots),
  plus ``None'' and ``Not sure.''

\item \textbf{Supporting Step.}
  ``Which step introduces an important supporting fact, rule, or
  verification?''
  Options: enumerated reasoning steps (Step~1, Step~2, \ldots),
  plus ``None'' and ``Not sure.''
\end{enumerate}

\smallskip\noindent\textit{D2 --- Error detection and localization (1--3 items).}

\begin{enumerate}[
    noitemsep,
    topsep=0.1em,
    leftmargin=1.2em
]
\item \textbf{Error Detection and Localization.}
  ``If you believe there is an error in this reasoning, which step
  most likely contains the incorrect reasoning?''
  Options: ``No error,'' enumerated reasoning steps (Step~1,
  Step~2, \ldots, up to Step~10), and ``I cannot determine.''

\item \textbf{Error Type.}
  ``What best describes the issue with this step?''
  \begin{itemize}
    \item Incorrect calculation
    \item Incorrect assumption
    \item Missing reasoning step
    \item Irrelevant reasoning
    \item Incorrect fact, rule, or formula
    \item Others (free-text)
  \end{itemize}
  \emph{Shown only if the participant selected a specific step in
  the localization item; skipped if ``No error'' was selected.}

\item \textbf{Error Impact.}
  ``If there is an error, does it affect the final answer?''
  \begin{itemize}
    \item Yes
    \item No
    \item Not sure
  \end{itemize}
  \emph{Shown only if the participant selected a specific step in
  the localization item; skipped if ``No error'' was selected.}
\end{enumerate}

\smallskip\noindent\textit{D3 --- Trust and reliance (3 items, 5-point Likert).}

\begin{enumerate}[
    noitemsep,
    topsep=0.1em,
    leftmargin=1.2em
]
\item \textbf{Reliance.}
  ``To what extent would you rely on this reasoning when making a
  real decision?''
  5-point scale:
  \begin{itemize}
    \item 1 -- I would not rely on it
    \item 2 -- I would probably not rely on it
    \item 3 -- Neutral
    \item 4 -- I would probably rely on it
    \item 5 -- I would rely on it
  \end{itemize}

\item \textbf{Verification Intention.}
  ``If you needed to make a real decision, would you verify this
  reasoning before using it?''
  5-point scale:
  \begin{itemize}
    \item 1 -- I would definitely verify it
    \item 2 -- I would probably verify it
    \item 3 -- I am not sure
    \item 4 -- I would probably not verify it
    \item 5 -- I would not verify it
  \end{itemize}
  Note: lower values indicate stronger verification intention.

\item \textbf{Interpretability.}
  ``How easy was it to understand this reasoning explanation?''
  5-point scale:
  \begin{itemize}
    \item 1 -- Very difficult
    \item 2 -- Difficult
    \item 3 -- Neutral
    \item 4 -- Easy
    \item 5 -- Very easy
  \end{itemize}
\end{enumerate}

\topic{Post-Survey.}
Participants answer the following items based on their experience
across the six reasoning representations shown in this study. For
each item, the participant selects one of the six representations
(presented as ``Reasoning format A''~through ``Reasoning format
F,'' with format labels mapped to representations in a randomized
order per participant) or ``No difference / Not sure.''
\smallskip\noindent\textit{Comprehension (3 items).}

\begin{enumerate}[
    noitemsep,
    topsep=0.1em,
    leftmargin=1.2em
]
\item \textbf{Step Comprehension.}
  ``Which explanation format helped you understand how the problem
  was solved most clearly?''

\item \textbf{Dependency Comprehension.}
  ``Which explanation format helped you best understand how
  different reasoning steps relate to each other?''

\item \textbf{Traceability.}
  ``Which explanation format made it easiest to trace how the
  final answer was derived from earlier reasoning steps?''
\end{enumerate}

\smallskip\noindent\textit{Error handling (2 items).}

\begin{enumerate}[
    noitemsep,
    topsep=0.1em,
    leftmargin=1.2em
]
\setcounter{enumi}{3}
\item \textbf{Error Detectability.}
  ``Which explanation format made it easiest to detect incorrect
  reasoning steps?''

\item \textbf{Error Localizability.}
  ``Which explanation format made it easiest to identify where the
  reasoning went wrong?''
\end{enumerate}

\smallskip\noindent\textit{Trust-related (3 items).}

\begin{enumerate}[
    noitemsep,
    topsep=0.1em,
    leftmargin=1.2em
]
\setcounter{enumi}{5}
\item \textbf{Reliability.}
  ``Which explanation format helped you best judge whether the
  model's reasoning was reliable?''

\item \textbf{Decision Confidence.}
  ``Which explanation format helped you feel most confident when
  deciding whether to trust the model's reasoning?''

\item \textbf{Critical Evaluation.}
  ``Which explanation format helped you critically evaluate the
  solution instead of accepting it without question?''
\end{enumerate}

\smallskip\noindent\textit{Overall preference (1 item).}

\begin{enumerate}[
    noitemsep,
    topsep=0.1em,
    leftmargin=1.2em
]
\setcounter{enumi}{8}
\item \textbf{Overall Usefulness.}
  ``Overall, which explanation format did you find most helpful
  for evaluating AI-generated outputs?''
\end{enumerate}

\subsection{Participants and Consent}
\label{app:irb}

\topic{Eligibility.}
Participants were required to be at least 18 years of age
and comfortable reading English.

\topic{Recruitment.}
Participants were recruited through in-class announcements
in courses taught by the principal investigator at the
authors' institution.
Interested students received a consent form confirming
the two eligibility requirements above, and all who
consented received the survey link; there was no
screening or selection beyond these requirements.
The pre-survey (six items; Appendix~\ref{app:survey})
was administered after consent and collected background
covariates only; no participant was excluded based on it.
In total, 64 participants began the survey.
Thirteen did not complete all phases and their partial
responses were excluded from analysis; 51 completed all
phases, and one of these completed after the
data-collection window closed, yielding the analytic
sample of $n = 50$.

\topic{Consent.}
Informed consent was obtained electronically through the
survey interface before any study data was collected.
Participants provided their name as an indication of consent;
names were stored separately from survey responses and were
not linked to any analytic data.

\topic{Compensation.}
Participants received compensation consistent with local
minimum wage guidelines as approved by the IRB, delivered
as an Amazon gift card. Participants who withdrew before
completing the session received prorated compensation based
on the portion of the study completed. To deliver the gift
card, participants provided an email address on a separate
form after completing the survey; the email address was
stored separately from survey responses.

\topic{Data handling.}
Responses were stored in anonymized form. No identifying
information beyond the consent name and gift-card delivery
email was collected, and neither was linked to participants'
analytic data. Per-item response data and per-item timing
are retained for analysis and will be released alongside
the paper.

\begin{table*}[t]
\centering
\small
\setlength{\tabcolsep}{5pt}
\renewcommand{\arraystretch}{1.15}
\begin{tabular}{lccccc}
\toprule
Representation & Sensitivity~$\uparrow$ & False-alarm~$\downarrow$ & Localization~$\uparrow$ & Type accuracy~$\uparrow$ & Impact-yes~$\uparrow$ \\
\midrule
Standard I/O       & 0.962 (25/26) & 0.125 (3/24)  & 0.600 (15/25) & 0.400 (6/15)  & 0.960 (24/25) \\
Zero-shot CoT      & \cellcolor{worstcell}0.917 (22/24) & \cellcolor{bestcell}\textbf{0.077 (2/26)}  & \cellcolor{bestcell}\textbf{0.955 (21/22)} & 0.476 (10/21) & 0.955 (21/22) \\
CoT-SC             & 0.935 (43/46) & 0.125 (13/104) & 0.732 (30/41) & 0.467 (14/30) & \cellcolor{bestcell}\textbf{1.000 (41/41)} \\
Plan-and-Solve     & 0.958 (23/24) & 0.192 (5/26)  & 0.571 (12/21) & \cellcolor{worstcell}0.000 (0/12)  & 0.952 (20/21) \\
Least-to-Most      & 0.923 (24/26) & \cellcolor{worstcell}0.208 (5/24)  & 0.905 (19/21) & \cellcolor{bestcell}\textbf{0.579 (11/19)} & \cellcolor{worstcell}0.905 (19/21) \\
Buffer of Thoughts & \cellcolor{bestcell}\textbf{0.963 (26/27)} & 0.174 (4/23)  & \cellcolor{worstcell}0.522 (12/23) & 0.167 (2/12)  & \cellcolor{bestcell}\textbf{1.000 (23/23)} \\
\bottomrule
\end{tabular}
\vspace{0.5em}
\begin{tabular}{lcccccrc}
\toprule
Representation & L0~missed & L1~noticed & L2~located & L3~diagnosed & L4~projected & $n$ & Mean \\
\midrule
Standard I/O       & 1  & 10 & 9  & 1 & 5  & 26 & 1.96 \\
Zero-shot CoT      & 2  & 1  & 11 & 0 & 10 & 24 & \cellcolor{bestcell}\textbf{2.62} \\
CoT-SC             & 3  & 13 & 16 & 0 & 14 & 46 & 2.20 \\
Plan-and-Solve     & 1  & 11 & 12 & 0 & 0  & 24 & \cellcolor{worstcell}1.46 \\
Least-to-Most      & 2  & 5  & 8  & 0 & 11 & 26 & 2.50 \\
Buffer of Thoughts & 1  & 14 & 10 & 0 & 2  & 27 & 1.56 \\
\bottomrule
\end{tabular}
\caption{D2 results across six representations. \textbf{Top panel}: \emph{Sensitivity} and \emph{false-alarm} are the proportion of errored (resp.\ correct) trials flagged with a step or ``cannot determine.'' \emph{Localization} is the proportion of step-flagged errored trials whose step set contains the injected step. \emph{Type accuracy} is conditional on correct localization, comparing user type against ground-truth type under the mapping \{calc$\to$calc, assumption/fact$\to$premise, missing$\to$missing, irrelevant$\to$irrelevant\}. \emph{Impact-yes} is the proportion of step-flagged errored trials where the user reported the error affects the final answer. \textbf{Bottom panel}: verification depth distribution on errored trials. Each trial is assigned to the deepest level reached: L0 (missed, no error flag), L1 (noticed, wrong step), L2 (located, wrong type), L3 (diagnosed, missing impact), L4 (projected, full verification). Mean depth $= \sum_{i=0}^{4} i \cdot n_i / n$.}
\label{tab:rq2_summary}
\vspace{-1.2em}
\end{table*}

\section{Additional Experimental Results}
\label{sec:additional_results}

We show the per-trace results that supplement the main
findings in \S\ref{sec:results}, one subsection per evaluation
dimension, followed by the post-survey results.

\subsection{Structural Understanding (D1)}

Table~\ref{tab:rq1_summary} in the main text reports the full D1
results: transition recognition and key-/support-step agreement
(Krippendorff's $\alpha$) for each representation.

\begin{table}[ht]
\centering
\small
\setlength{\tabcolsep}{6pt}
\renewcommand{\arraystretch}{1.1}
\begin{tabular}{lcc}
\toprule
\textbf{Representation} & \textbf{Count} & \textbf{\%} \\
\midrule
\cellcolor{winnerGreen}Plan-and-Solve (P\&S)   & \cellcolor{winnerGreen}13 & \cellcolor{winnerGreen}26.0 \\
Zero-shot CoT (ZS-CoT)                         & 11 & 22.0 \\
Least-to-Most (L2M)                            & 11 & 22.0 \\
Buffer of Thoughts (BoT)                       & 7  & 14.0 \\
No difference (NoDiff)                         & 5  & 10.0 \\
Standard I/O (SI)                              & 3  & 6.0  \\
CoT-Self-Consistency (CoT-SC)                  & 0  & 0.0  \\
\midrule
\textbf{Total}                                 & \textbf{50} & \textbf{100.0} \\
\bottomrule
\end{tabular}
\caption{Overall usefulness preference: representation chosen as ``most helpful overall'' by each completed participant. The modal choice is highlighted in green.}
\label{tab:overall_preference}
\vspace{-1.2em}
\end{table}

\subsection{Verification Performance (D2)}

Table~\ref{tab:rq2_summary} shows the full D2 results.

\begin{table}[ht]
\centering
\footnotesize
\setlength{\tabcolsep}{3pt}
\renewcommand{\arraystretch}{1.15}
\begin{tabular}{lccc}
\toprule
Representation & Rel.~$\uparrow$ & Verif.~$\downarrow$ & Interp.~$\uparrow$ \\
\midrule
Standard I/O       & \cellcolor{worstcell}3.32 (1.25) & \cellcolor{bestcell}\textbf{1.84 (1.15)} & 3.66 (1.22) \\
Zero-shot CoT      & \cellcolor{bestcell}\textbf{4.04 (1.11)} & 2.44 (1.36) & \cellcolor{bestcell}\textbf{4.12 (1.00)} \\
CoT-SC             & 3.37 (1.39) & 2.30 (1.42) & 3.61 (1.32) \\
Plan-and-Solve     & 3.88 (1.22) & \cellcolor{worstcell}2.64 (1.54) & 3.62 (1.23) \\
Least-to-Most      & 3.52 (1.22) & 2.46 (1.34) & 3.28 (1.14) \\
Buffer of Thoughts & 3.54 (1.34) & 2.48 (1.39) & \cellcolor{worstcell}3.08 (1.29) \\
\bottomrule
\end{tabular}
\caption{D3 results across six representations. Mean (standard deviation) of 5-point Likert responses per (participant, representation). \emph{Rel.} (reliance): ``To what extent would you rely on this reasoning when making a decision?'' (1 = would not rely, 5 = would rely). \emph{Verif.} (verification intention): ``Would you verify this reasoning before using it?'' (1 = would definitely verify, 5 = would not verify); lower indicates stronger verification intention. \emph{Interp.} (interpretability): ``How easy was it to understand this reasoning?'' (1 = very difficult, 5 = very easy). $n=50$ per representation except CoT-SC ($n=150$ due to per-path responses). Green marks the best value per column; red marks the worst.}
\label{tab:rq3_summary}
\vspace{-1.2em}
\end{table}

\subsection{Trust Calibration (D3)}

Table~\ref{tab:rq3_summary} reports the mean and standard deviation
of the three trust items for each representation.

\subsection{Post-Survey Preference}

Table~\ref{tab:post_survey_counts} in the main text reports
item-level selection counts for the eight post-survey items, and
Table~\ref{tab:overall_preference} reports overall preference
counts.

\section{AI Assistance Disclosure}
\label{app:use-of-ai}

In preparing this work, the authors used OpenAI's ChatGPT
for limited assistance with manuscript preparation
(grammar correction and stylistic refinement)
in accordance with ACL policies on AI-assisted writing.
All AI-assisted text was reviewed, edited, and verified
by the authors, who remain responsible for the
final content of the paper.

\end{document}